\documentclass{article}

\PassOptionsToPackage{numbers,sort}{natbib}
\usepackage[preprint]{neurips_2021}

\usepackage[utf8]{inputenc} % allow utf-8 input
\usepackage[T1]{fontenc}    % use 8-bit T1 fonts
\usepackage{hyperref}       % hyperlinks
\usepackage{url}            % simple URL typesetting
\usepackage{booktabs}       % professional-quality tables
\usepackage{amsfonts}       % blackboard math symbols
\usepackage{nicefrac}       % compact symbols for 1/2, etc.
\usepackage{microtype}      % microtypography
\usepackage{xcolor}         % colors

\usepackage[american]{babel}
\usepackage[nolist]{acronym}
\usepackage{amsmath,amssymb}
\usepackage{pgfplots}
\DeclareUnicodeCharacter{2212}{−}
\usepgfplotslibrary{groupplots,dateplot}
\pgfplotsset{compat=newest,}
\usepackage{subcaption}
\usepackage{adjustbox}
\usepackage{xcolor}

\definecolor{code-green}{RGB}{0, 204, 153}
\newcommand{\code}[2]{\textcolor{#1}{\texttt{#2}}}

\newcommand{\citef}[1]{{\tiny\citet{#1}}}

\newcommand\modelCount{19}
\newcommand\LossFunc{Jitter}

\usetikzlibrary{math,patterns}
\tikzmath
{
  function symlog(\x,\a){
    \yLarge = ((\x>\a) - (\x<-\a)) * (ln(max(abs(\x/\a),1)) + 1);
    \ySmall = (\x >= -\a) * (\x <= \a) * \x / \a ;
    return \yLarge + \ySmall ;
  };
  function symexp(\y,\a){
    \xLarge = ((\y>1) - (\y<-1)) * \a * exp(abs(\y) - 1) ;
    \xSmall = (\y>=-1) * (\y<=1) * \a * \y ;
    return \xLarge + \xSmall ;
  };
}

\let\pgfimageWithoutPath\pgfimage 
\renewcommand{\pgfimage}[2][]{\pgfimageWithoutPath[#1]{Images/TikzImages/#2}}
\title{Exploring Misclassifications of Robust Neural Networks to Enhance Adversarial Attacks}

\author{%
Leo Schwinn, René Raab, An Nguyen, Dario Zanca, Bjoern Eskofier \\
Department Artificial Intelligence in Biomedical Engineering,
    Univ. of Erlangen-Nürnberg,
    Germany \\
    \texttt{\{leo.schwinn,rene.raab,an.nguyen,dario.zanca,bjoern.eskofier\}@fau.de}

}

\begin{document}

\begin{acronym}
\acrodef{DNN}{Deep Neural Network}
\acrodef{DLR}{Difference of Logit Ratio}
\acrodef{CE}{Cross-Entropy}
\acrodef{CW}{Carlini \& Wagner}
\acrodef{PGD}{Projected Gradient Descent}
\acro{APGD}{Auto-PGD}
\end{acronym}

\maketitle

\begin{abstract}
Progress in making neural networks more robust against adversarial attacks is mostly marginal, despite the great efforts of the research community. Moreover, the robustness evaluation is often imprecise, making it difficult to identify promising approaches. We analyze the classification decisions of $19$ different state-of-the-art neural networks trained to be robust against adversarial attacks. Our findings suggest that current untargeted adversarial attacks induce misclassification towards only a limited amount of different classes. Additionally, we observe that both over- and under-confidence in model predictions result in an inaccurate assessment of model robustness. Based on these observations, we propose a novel loss function for adversarial attacks that consistently improves attack success rate compared to prior loss functions for $19$ out of $19$ analyzed models.

%Based on these observations, we design a novel loss function that encourages diversity of attacked classes and is invariant to extreme confidence values. The proposed loss function consistently improves attack success rate compared to prior loss functions for $19$ out of $19$ different models from the literature.
\end{abstract}

\section{Introduction}

%Standard Intro
\acfp{DNN} can be easily fooled into making wrong predictions by seemingly negligible perturbations to their input data, called adversarial examples. \citet{Szegedy14} first demonstrated the existence of adversarial examples for neural networks in the image domain. Since then, adversarial examples have been identified in various other domains such as speech recognition \cite{Qin2019Imperceptible} and natural language processing \cite{Morris20Textattack}. This prevalence of adversarial examples has severe security implications for real-world applications. As a result, the robustness of neural networks to adversarial examples has become a central research topic of deep learning in recent years. 

%Problems of evaluating robustness
Several defense strategies have been proposed to make \acp{DNN} more robust~\cite{Carmon2019Unlabeld, Ding2020MMA, Hendrycks2019Using, Madry2018Adversarial}. However, most of them have later been shown to be ineffective against stronger attacks~\cite{Jin2020Manifold,Pang2020Boosting} and overall progress has been slow \cite{croce2020reliable}. As robustness improvements are mostly marginal, a reliable evaluation of new defense strategies is critical to identify methods that actually improve robustness. Therefore, the community has established helpful guidelines to reliably evaluate new defenses \cite{Athalye2018Obfuscated, Tramer2020, Useato2018}. Nevertheless, the worst-case robustness of \acp{DNN} is still reduced repetitively by even stronger attacks and a precise evaluation remains a challenging problem \cite{croce2020reliable}. Moreover, prior work focuses on evaluating the robustness of individual \acp{DNN} without bringing them into the context of other models.

%Our Work
In this work, we explore the classification decisions of $\modelCount$ recently published \acp{DNN}. Hereby, we restrict our analysis to \acp{DNN} which have been trained to be robust against adversarial attacks with a variety of different methods. Our analysis can be summarized by two main findings: First, we observe that untargeted adversarial attacks cause misclassification towards only a limited amount of different classes in the dataset. Second, we identify three model properties that make it difficult to accurately assess model robustness -- namely, model over- and under-confidence, large output logits, and irregular input gradient geometry. We leverage these observations to design a new loss function that improves the success rate of adversarial attacks compared to current state-of-the-art loss functions. More specifically, we encourage attack diversity in untargeted attacks by injecting noise to the model output. Additionally, we introduce scale invariance to the loss function by normalizing the output logits to a fixed value range. Thereby, we circumvent the gradient obfuscation problem generated by models with low-confidence predictions or irregularly large output logits \cite{Carlini017Towards,croce2020reliable}.  
%solve the gradient obfuscation problem generated  and additionally by a small standard deviation between logits that leads to low confidence predictions as shown in our analysis. 
Moreover, we propose a simple yet effective mechanism that minimizes the magnitude of perturbations, as shown in Figure~\ref{fig:perturbed_images}, without compromising the success rate of an attack. This leads to the definition of an objective function for adversarial attacks, which we will refer to as \textit{\LossFunc{}}. We empirically evaluate our loss function on an extensive benchmark consisting of $\modelCount$ different models proposed in the literature. We show that \LossFunc{}-based attacks consistently improve the success rate compared to prior loss functions in all $\modelCount$ models by up to $13.8$ percentage points. Additionally, \LossFunc{}-based attacks generate perturbations with a $2.85$ times smaller norm on average. Lastly, we analyze the effect of \LossFunc{} on the classification decisions to explain its effectiveness.

%Second, we discover that the robustness of models which show an exceptionally small standard deviation between their output logits is more difficult to evaluate.

{\def\figS{3.1cm}
\begin{figure}
    \centering
    \input{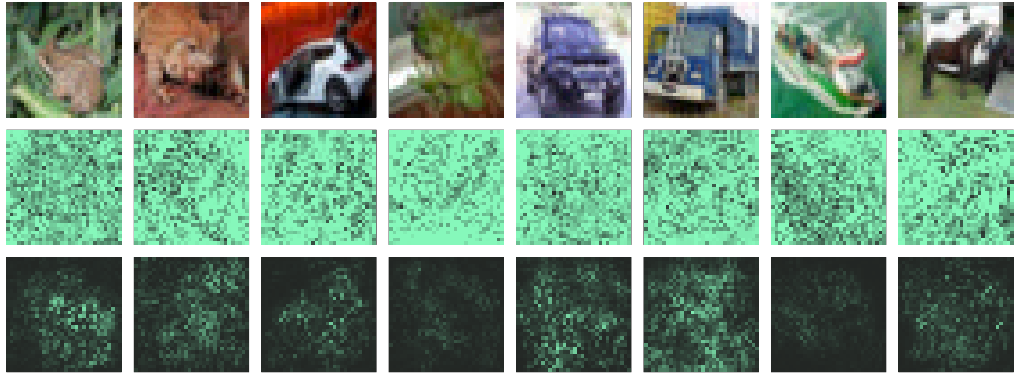}
    \caption{Difference of adversarial perturbations created by \ac{CE}-based attacks and \LossFunc-based attacks. Original images are shown in the first row, CE-based perturbations in the second row, and \LossFunc{}-based perturbations in the last row.}
    \label{fig:perturbed_images}
\end{figure}
}

%\section{Preliminaries}

%We first introduce the mathematical notation used in this work. We then cover recent advances in the field of adversarial attacks.

\section{Notation}

%Mathematical notation
Let  $f_{\theta}: [0, 1]^d \rightarrow \mathbb{R}^C$ be a \ac{DNN} classifier parameterized by $\theta \in \Theta$ with $f_{\theta}: x \mapsto z$. Here $x$ is a $d$-dimensional input image, $z$ is the respective output vector (logits) of the \ac{DNN}, and $C$ denotes the number of classes. The ground truth class label of a given image is described by $y \in \{1, \dots, C\}$ while the predicted class label $\hat{y} \in \{1, \dots, C\}$ is given by $\operatorname{argmax(z)}$. 

%Adversarial attack definition
Adversarial examples $x_{adv} = x + \gamma$ aim to change the input data of \acp{DNN} such that the classification decision of the network is altered, but the class label remains the same for human perception. Additionally, $x_{adv}$ is restricted to remain within the data domain, i.e. $x_{adv} \in [0, 1]^d$. A common way to enforce semantic similarity to the original sample is to restrict the magnitude $\epsilon$ of the adversarial perturbation $\gamma$ by a $\ell_p$-norm bound, such that $||\gamma||_p \leq \epsilon$. We refer to the set of valid adversarial examples that fulfill these constraints as $S$. As prior work mainly focuses on $p = \infty$ and thus most models are available for this threat model, we focus on $p = \infty$ in this work as well. Furthermore, we restrict our analysis to untargeted white-box adversarial attacks.

%Finding an adversarial attack can then be formulated as the following optimization problem: 

%\begin{align} \label{eq:adversarial_opt}
%\underset{(x + \gamma) \in S}\max ~\mathcal{L}(f_{\theta}(x + \gamma), y),
%\end{align}

\section{Related work}

One of the most often used adversarial attacks, \acf{PGD}, was proposed by \citet{Madry2018Adversarial}. \ac{PGD} is an iterative gradient-based attack, in which multiple smaller gradient updates are used to find the adversarial perturbation:
\begin{align} \label{eq:ifgsm}
    x_{adv}^{t + 1} = \Pi_{S} \, (x_{adv}^{t} + \alpha \cdot \operatorname{sign}(\nabla_x  \mathcal{L}(f_\theta(x_{adv}^{t}), y))
\end{align}

where $0 < \alpha \leq \epsilon$ and $x_{adv}^{t}$ describes the adversarial example at iteration $t$. The loss of the attack is given by $\mathcal{L}(f_{\theta}(x_{adv}^t), y)$. $\Pi_{S}(x)$ is a projection operator that keeps $\gamma^t$ within the set of valid perturbations $S$ and $\operatorname{sign}$ is the component-wise signum operator. The starting point of the attack $x_{adv}^0$ is randomly chosen in the $\epsilon$-norm ball. More variants of iterative gradient-based attacks have been proposed that are more effective than vanilla \ac{PGD} \cite{Lin2020, Schwinn2021dynamically, Useato2018}. Recently, \citet{croce2020reliable} proposed the \acf{APGD} attack. In contrast to previous \ac{PGD} versions, \ac{APGD} requires considerably less hyperparameter tuning and was shown to be more effective than other \ac{PGD}-based attacks against a variety of models \cite{croce2020reliable}. Nevertheless, one important component of all gradient-based attacks is their optimization objective. The most often used objective is the \acf{CE} loss. \citet{Carlini017Towards} observe that \ac{CE}-based attacks fail against models with large logits. They propose the \ac{CW} loss function $-z_y + \operatorname{max}_{i \neq y} (z_i)$ which does not make use of the softmax function and thereby reduces the scaling problem. Nevertheless, \citet{croce2020reliable} observe that the scale dependence of the \ac{CW} loss can still lead to severe failure cases against models with exceptionally large logits. They address this issue with the scale- and shift-invariant \acf{DLR} loss and show its effectiveness on an extensive benchmark.

%Other advances in the field of adversarial attacks include a diverse initialization of the starting perturbation \cite{tashiro2020} or combining multiple different attack algorithms in a single attack \cite{Mao2020}. Recently, \citet{croce2020reliable} proposed a sophisticated attack combination that efficiently circumvents the most common evaluation flaws to reduce the amount of manual tuning and expert knowledge required to reliably evaluate new defenses. Nevertheless, a correct evaluation is still complicated and expensive.

\section{Robust misclassifications}

In this section we explore the classification decisions of $\modelCount$ different models in the presence of adversarial attacks. We restrict our analysis to models trained on the CIFAR10 dataset.\footnote{The labels "airplane" and "automobile" have been changed to "plane" and "car", respectively} We choose the recently proposed \acf{APGD} with the \acf{DLR} loss as an attack to perturb the images, as it is one of the most efficient and reliable gradient-based attacks \cite{croce2020reliable}. These choices and specific hyperparameters are described in more detail in Section \ref{sec:experiments}.

\subsection{Distribution of misclassifications}

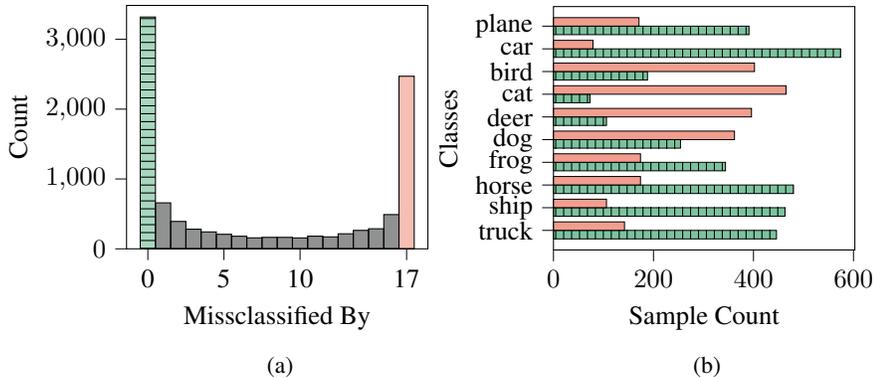
\begin{figure}
    \centering
    \begin{subfigure}[b]{0.4\textwidth}
        \centering
        % This file was created by tikzplotlib v0.9.8.
\begin{tikzpicture}[
trim axis left, trim axis right
]

\definecolor{color0}{rgb}{0.311718304369156,0.68826334252299,0.49429685756019}
\definecolor{color1}{rgb}{0.134998800024609,0.142173014868963,0.137714103944917}
\definecolor{color2}{rgb}{0.929372737926104,0.495868206115888,0.409870571976527}

\begin{axis}[
tick align=outside,
tick pos=left,
width=\linewidth,
x grid style={white!69.0196078431373!black},
xlabel={Missclassified By},
xmin=-0.9, xmax=18.9,
xtick style={color=black},
xtick={0.5, 5.5, 10.5,17.5},
xticklabels={0, 5, 10, 17},
y grid style={white!69.0196078431373!black},
ylabel={Count},
ymin=0, ymax=3484.95,
ytick style={color=black},
]
\draw[draw=black,fill=color0,fill opacity=0.5,postaction={pattern=horizontal lines}] (axis cs:0,0) rectangle (axis cs:1,3319);
\draw[draw=black,fill=color0,fill opacity=0.5] (axis cs:1,0) rectangle (axis cs:2,0);
\draw[draw=black,fill=color0,fill opacity=0.5] (axis cs:2,0) rectangle (axis cs:3,0);
\draw[draw=black,fill=color0,fill opacity=0.5] (axis cs:3,0) rectangle (axis cs:4,0);
\draw[draw=black,fill=color0,fill opacity=0.5] (axis cs:4,0) rectangle (axis cs:5,0);
\draw[draw=black,fill=color0,fill opacity=0.5] (axis cs:5,0) rectangle (axis cs:6,0);
\draw[draw=black,fill=color0,fill opacity=0.5] (axis cs:6,0) rectangle (axis cs:7,0);
\draw[draw=black,fill=color0,fill opacity=0.5] (axis cs:7,0) rectangle (axis cs:8,0);
\draw[draw=black,fill=color0,fill opacity=0.5] (axis cs:8,0) rectangle (axis cs:9,0);
\draw[draw=black,fill=color0,fill opacity=0.5] (axis cs:9,0) rectangle (axis cs:10,0);
\draw[draw=black,fill=color0,fill opacity=0.5] (axis cs:10,0) rectangle (axis cs:11,0);
\draw[draw=black,fill=color0,fill opacity=0.5] (axis cs:11,0) rectangle (axis cs:12,0);
\draw[draw=black,fill=color0,fill opacity=0.5] (axis cs:12,0) rectangle (axis cs:13,0);
\draw[draw=black,fill=color0,fill opacity=0.5] (axis cs:13,0) rectangle (axis cs:14,0);
\draw[draw=black,fill=color0,fill opacity=0.5] (axis cs:14,0) rectangle (axis cs:15,0);
\draw[draw=black,fill=color0,fill opacity=0.5] (axis cs:15,0) rectangle (axis cs:16,0);
\draw[draw=black,fill=color0,fill opacity=0.5] (axis cs:16,0) rectangle (axis cs:17,0);
\draw[draw=black,fill=color0,fill opacity=0.5] (axis cs:17,0) rectangle (axis cs:18,0);
\draw[draw=black,fill=color1,fill opacity=0.5] (axis cs:0,0) rectangle (axis cs:1,0);
\draw[draw=black,fill=color1,fill opacity=0.5] (axis cs:1,0) rectangle (axis cs:2,658);
\draw[draw=black,fill=color1,fill opacity=0.5] (axis cs:2,0) rectangle (axis cs:3,392);
\draw[draw=black,fill=color1,fill opacity=0.5] (axis cs:3,0) rectangle (axis cs:4,281);
\draw[draw=black,fill=color1,fill opacity=0.5] (axis cs:4,0) rectangle (axis cs:5,242);
\draw[draw=black,fill=color1,fill opacity=0.5] (axis cs:5,0) rectangle (axis cs:6,208);
\draw[draw=black,fill=color1,fill opacity=0.5] (axis cs:6,0) rectangle (axis cs:7,180);
\draw[draw=black,fill=color1,fill opacity=0.5] (axis cs:7,0) rectangle (axis cs:8,157);
\draw[draw=black,fill=color1,fill opacity=0.5] (axis cs:8,0) rectangle (axis cs:9,164);
\draw[draw=black,fill=color1,fill opacity=0.5] (axis cs:9,0) rectangle (axis cs:10,164);
\draw[draw=black,fill=color1,fill opacity=0.5] (axis cs:10,0) rectangle (axis cs:11,156);
\draw[draw=black,fill=color1,fill opacity=0.5] (axis cs:11,0) rectangle (axis cs:12,181);
\draw[draw=black,fill=color1,fill opacity=0.5] (axis cs:12,0) rectangle (axis cs:13,171);
\draw[draw=black,fill=color1,fill opacity=0.5] (axis cs:13,0) rectangle (axis cs:14,215);
\draw[draw=black,fill=color1,fill opacity=0.5] (axis cs:14,0) rectangle (axis cs:15,264);
\draw[draw=black,fill=color1,fill opacity=0.5] (axis cs:15,0) rectangle (axis cs:16,287);
\draw[draw=black,fill=color1,fill opacity=0.5] (axis cs:16,0) rectangle (axis cs:17,490);
\draw[draw=black,fill=color1,fill opacity=0.5] (axis cs:17,0) rectangle (axis cs:18,0);
\draw[draw=black,fill=color2,fill opacity=0.5] (axis cs:0,0) rectangle (axis cs:1,0);
\draw[draw=black,fill=color2,fill opacity=0.5] (axis cs:1,0) rectangle (axis cs:2,0);
\draw[draw=black,fill=color2,fill opacity=0.5] (axis cs:2,0) rectangle (axis cs:3,0);
\draw[draw=black,fill=color2,fill opacity=0.5] (axis cs:3,0) rectangle (axis cs:4,0);
\draw[draw=black,fill=color2,fill opacity=0.5] (axis cs:4,0) rectangle (axis cs:5,0);
\draw[draw=black,fill=color2,fill opacity=0.5] (axis cs:5,0) rectangle (axis cs:6,0);
\draw[draw=black,fill=color2,fill opacity=0.5] (axis cs:6,0) rectangle (axis cs:7,0);
\draw[draw=black,fill=color2,fill opacity=0.5] (axis cs:7,0) rectangle (axis cs:8,0);
\draw[draw=black,fill=color2,fill opacity=0.5] (axis cs:8,0) rectangle (axis cs:9,0);
\draw[draw=black,fill=color2,fill opacity=0.5] (axis cs:9,0) rectangle (axis cs:10,0);
\draw[draw=black,fill=color2,fill opacity=0.5] (axis cs:10,0) rectangle (axis cs:11,0);
\draw[draw=black,fill=color2,fill opacity=0.5] (axis cs:11,0) rectangle (axis cs:12,0);
\draw[draw=black,fill=color2,fill opacity=0.5] (axis cs:12,0) rectangle (axis cs:13,0);
\draw[draw=black,fill=color2,fill opacity=0.5] (axis cs:13,0) rectangle (axis cs:14,0);
\draw[draw=black,fill=color2,fill opacity=0.5] (axis cs:14,0) rectangle (axis cs:15,0);
\draw[draw=black,fill=color2,fill opacity=0.5] (axis cs:15,0) rectangle (axis cs:16,0);
\draw[draw=black,fill=color2,fill opacity=0.5] (axis cs:16,0) rectangle (axis cs:17,0);
\draw[draw=black,fill=color2,fill opacity=0.5] (axis cs:17,0) rectangle (axis cs:18,2471);
\end{axis}

\end{tikzpicture}
        \caption{}
        \label{fig:distribution_misclassification_a}
    \end{subfigure}
    \begin{subfigure}[b]{0.4\textwidth}
        \centering
        % This file was created by tikzplotlib v0.9.8.
\begin{tikzpicture}[
trim axis left, trim axis right
]

\definecolor{color0}{rgb}{0.311718304369156,0.68826334252299,0.49429685756019}
\definecolor{color1}{rgb}{0.929372737926104,0.495868206115888,0.409870571976527}

\begin{axis}[
legend style={fill opacity=0.8, draw opacity=1, text opacity=1, draw=none},
tick align=outside,
tick pos=left,
width=\linewidth,
x grid style={white!69.0196078431373!black},
xlabel={Sample Count},
xmin=0, xmax=602.7,
xtick style={color=black},
y dir=reverse,
y grid style={white!69.0196078431373!black},
ylabel={Classes},
ymin=-0.8625, ymax=9.8625,
ytick style={color=black},
ytick={0,1,2,3,4,5,6,7,8,9},
yticklabels={plane,car,bird,cat,deer,dog,frog,horse,ship,truck}
]
\draw[draw=black,fill=color0,fill opacity=0.75,line width=-0.372184615384615pt,postaction={pattern=vertical lines}] (axis cs:0,0) rectangle (axis cs:391,0.375);
\draw[draw=black,fill=color0,fill opacity=0.75,line width=-0.372184615384615pt,postaction={pattern=vertical lines}] (axis cs:0,1) rectangle (axis cs:574,1.375);
\draw[draw=black,fill=color0,fill opacity=0.75,line width=-0.372184615384615pt,postaction={pattern=vertical lines}] (axis cs:0,2) rectangle (axis cs:188,2.375);
\draw[draw=black,fill=color0,fill opacity=0.75,line width=-0.372184615384615pt,postaction={pattern=vertical lines}] (axis cs:0,3) rectangle (axis cs:73,3.375);
\draw[draw=black,fill=color0,fill opacity=0.75,line width=-0.372184615384615pt,postaction={pattern=vertical lines}] (axis cs:0,4) rectangle (axis cs:106,4.375);
\draw[draw=black,fill=color0,fill opacity=0.75,line width=-0.372184615384615pt,postaction={pattern=vertical lines}] (axis cs:0,5) rectangle (axis cs:254,5.375);
\draw[draw=black,fill=color0,fill opacity=0.75,line width=-0.372184615384615pt,postaction={pattern=vertical lines}] (axis cs:0,6) rectangle (axis cs:344,6.375);
\draw[draw=black,fill=color0,fill opacity=0.75,line width=-0.372184615384615pt,postaction={pattern=vertical lines}] (axis cs:0,7) rectangle (axis cs:480,7.375);
\draw[draw=black,fill=color0,fill opacity=0.75,line width=-0.372184615384615pt,postaction={pattern=vertical lines}] (axis cs:0,8) rectangle (axis cs:463,8.375);
\draw[draw=black,fill=color0,fill opacity=0.75,line width=-0.372184615384615pt,postaction={pattern=vertical lines}] (axis cs:0,9) rectangle (axis cs:446,9.375);
\draw[draw=black,fill=color1,fill opacity=0.75,line width=-0.372184615384615pt] (axis cs:0,-0.375) rectangle (axis cs:171,0);
\draw[draw=black,fill=color1,fill opacity=0.75,line width=-0.372184615384615pt] (axis cs:0,0.625) rectangle (axis cs:79,1);
\draw[draw=black,fill=color1,fill opacity=0.75,line width=-0.372184615384615pt] (axis cs:0,1.625) rectangle (axis cs:402,2);
\draw[draw=black,fill=color1,fill opacity=0.75,line width=-0.372184615384615pt] (axis cs:0,2.625) rectangle (axis cs:465,3);
\draw[draw=black,fill=color1,fill opacity=0.75,line width=-0.372184615384615pt] (axis cs:0,3.625) rectangle (axis cs:396,4);
\draw[draw=black,fill=color1,fill opacity=0.75,line width=-0.372184615384615pt] (axis cs:0,4.625) rectangle (axis cs:362,5);
\draw[draw=black,fill=color1,fill opacity=0.75,line width=-0.372184615384615pt] (axis cs:0,5.625) rectangle (axis cs:174,6);
\draw[draw=black,fill=color1,fill opacity=0.75,line width=-0.372184615384615pt] (axis cs:0,6.625) rectangle (axis cs:174,7);
\draw[draw=black,fill=color1,fill opacity=0.75,line width=-0.372184615384615pt] (axis cs:0,7.625) rectangle (axis cs:106,8);
\draw[draw=black,fill=color1,fill opacity=0.75,line width=-0.372184615384615pt] (axis cs:0,8.625) rectangle (axis cs:142,9);
\end{axis}

\end{tikzpicture}
        \caption{}
        \label{fig:distribution_misclassification_b}
    \end{subfigure}
    \caption{Analysis of misclassification decisions for $17$ different models. Subfigure (a) shows by how many models each attacked input is misclassified. Robust images that are never misclassified are shown in the leftmost column (green, dashed) and non-robust images that are always misclassified are shown in the rightmost column (red). Subfigure (b) displays the difference between the class distributions between robust and non-robust images. Both statistics are calculated on the test set of CIFAR10.}
    \label{fig:distribution_misclassification}
\end{figure}

%description of model ensemble and 

Recent studies mainly focus on common evaluation metrics to assess the robustness of \acp{DNN}. This includes the worst-case robustness of a classifier \cite{croce2020reliable} and the magnitude of the perturbation norm necessary to fool the classifier for individual inputs \cite{Carlini017Towards}. Here, we provide insights into the classification decisions and numerical properties of a large and diverse set of models from the literature. We focus on models that are trained to be robust to adversarial attacks. Furthermore, all models are attacked individually to find the respective worst-case robustness.

%Missclassification distribution
Figure \ref{fig:distribution_misclassification_a} shows how the $17$ most robust models misclassify inputs attacked by \ac{APGD}. We left out the models by \citet{Jin2020Manifold} and \citet{Mustafa2019Adversarial} from this analysis, as they show negligible robustness against strong adversarial attacks. Out of the $10,000$ test samples of the CIFAR10 dataset, $3319$ are correctly classified by all $17$ models, while $2471$ samples are consistently misclassified by all models. This is shown by the leftmost (green, dashed) and rightmost (red) bar of the histogram plot. Inspired by prior work \cite{Ilyas2019Adversarial}, we will refer to images in the first group that are never misclassified as \emph{robust images} and images in the second group that are always misclassified as \emph{non-robust images}. The gray bars in between show the remaining $4210$ samples misclassified by at least one model but not by all models.
%Class distribution between robust and non-robust images
Figure \ref{fig:distribution_misclassification_b} summarizes the class distribution of robust and non-robust images. There is a considerable difference in frequency for most classes between the two groups. Images from the classes "plane", "car", "horse", "ship", and "truck" are often classified correctly while "bird", "cat", "deer" and "dog" are mostly misclassified. 

\begin{figure}
    \centering
    \begin{subfigure}[b]{0.42\textwidth}
        \centering
        % This file was created by tikzplotlib v0.9.8.
\begin{tikzpicture}[trim axis left, trim axis right]

\begin{axis}[
tick align=outside,
tick pos=left,
width=\linewidth,
x grid style={white!69.0196078431373!black},
xlabel={Adversarial Pred.},
xmin=0, xmax=10,
xtick style={color=black},
xtick={0.5,1.5,2.5,3.5,4.5,5.5,6.5,7.5,8.5,9.5},
xticklabel style={rotate=90.0},
xticklabels={plane,car,bird,cat,deer,dog,frog,horse,ship,truck},
y dir=reverse,
y grid style={white!69.0196078431373!black},
ylabel={Original Pred.},
ymin=0, ymax=10,
ytick style={color=black},
ytick={0.5,1.5,2.5,3.5,4.5,5.5,6.5,7.5,8.5,9.5},
yticklabels={plane,car,bird,cat,deer,dog,frog,horse,ship,truck}
]
\addplot graphics [includegraphics cmd=\pgfimage,xmin=0, xmax=10, ymin=10, ymax=0] {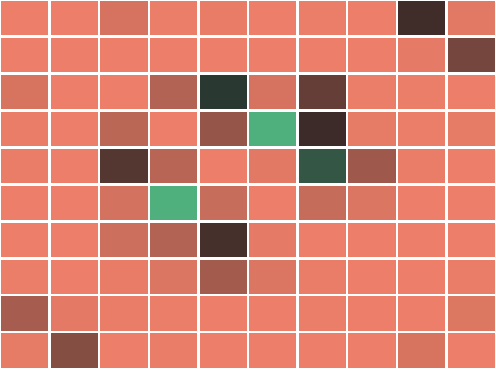};
\draw (axis cs:0.5,0.5) node[
  scale=0.6,
  text=white,
  rotate=0.0
]{0};
\draw (axis cs:1.5,0.5) node[
  scale=0.6,
  text=white,
  rotate=0.0
]{11};
\draw (axis cs:2.5,0.5) node[
  scale=0.6,
  text=white,
  rotate=0.0
]{61};
\draw (axis cs:3.5,0.5) node[
  scale=0.6,
  text=white,
  rotate=0.0
]{21};
\draw (axis cs:4.5,0.5) node[
  scale=0.6,
  text=white,
  rotate=0.0
]{29};
\draw (axis cs:5.5,0.5) node[
  scale=0.6,
  text=white,
  rotate=0.0
]{6};
\draw (axis cs:6.5,0.5) node[
  scale=0.6,
  text=white,
  rotate=0.0
]{20};
\draw (axis cs:7.5,0.5) node[
  scale=0.6,
  text=white,
  rotate=0.0
]{10};
\draw (axis cs:8.5,0.5) node[
  scale=0.6,
  text=white,
  rotate=0.0
]{162};
\draw (axis cs:9.5,0.5) node[
  scale=0.6,
  text=white,
  rotate=0.0
]{42};
\draw (axis cs:0.5,1.5) node[
  scale=0.6,
  text=white,
  rotate=0.0
]{13};
\draw (axis cs:1.5,1.5) node[
  scale=0.6,
  text=white,
  rotate=0.0
]{0};
\draw (axis cs:2.5,1.5) node[
  scale=0.6,
  text=white,
  rotate=0.0
]{7};
\draw (axis cs:3.5,1.5) node[
  scale=0.6,
  text=white,
  rotate=0.0
]{12};
\draw (axis cs:4.5,1.5) node[
  scale=0.6,
  text=white,
  rotate=0.0
]{4};
\draw (axis cs:5.5,1.5) node[
  scale=0.6,
  text=white,
  rotate=0.0
]{5};
\draw (axis cs:6.5,1.5) node[
  scale=0.6,
  text=white,
  rotate=0.0
]{12};
\draw (axis cs:7.5,1.5) node[
  scale=0.6,
  text=white,
  rotate=0.0
]{1};
\draw (axis cs:8.5,1.5) node[
  scale=0.6,
  text=white,
  rotate=0.0
]{36};
\draw (axis cs:9.5,1.5) node[
  scale=0.6,
  text=white,
  rotate=0.0
]{135};
\draw (axis cs:0.5,2.5) node[
  scale=0.6,
  text=white,
  rotate=0.0
]{59};
\draw (axis cs:1.5,2.5) node[
  scale=0.6,
  text=white,
  rotate=0.0
]{3};
\draw (axis cs:2.5,2.5) node[
  scale=0.6,
  text=white,
  rotate=0.0
]{0};
\draw (axis cs:3.5,2.5) node[
  scale=0.6,
  text=white,
  rotate=0.0
]{96};
\draw (axis cs:4.5,2.5) node[
  scale=0.6,
  text=white,
  rotate=0.0
]{184};
\draw (axis cs:5.5,2.5) node[
  scale=0.6,
  text=white,
  rotate=0.0
]{62};
\draw (axis cs:6.5,2.5) node[
  scale=0.6,
  text=white,
  rotate=0.0
]{144};
\draw (axis cs:7.5,2.5) node[
  scale=0.6,
  text=white,
  rotate=0.0
]{21};
\draw (axis cs:8.5,2.5) node[
  scale=0.6,
  text=white,
  rotate=0.0
]{16};
\draw (axis cs:9.5,2.5) node[
  scale=0.6,
  text=white,
  rotate=0.0
]{11};
\draw (axis cs:0.5,3.5) node[
  scale=0.6,
  text=white,
  rotate=0.0
]{23};
\draw (axis cs:1.5,3.5) node[
  scale=0.6,
  text=white,
  rotate=0.0
]{8};
\draw (axis cs:2.5,3.5) node[
  scale=0.6,
  text=white,
  rotate=0.0
]{88};
\draw (axis cs:3.5,3.5) node[
  scale=0.6,
  text=white,
  rotate=0.0
]{0};
\draw (axis cs:4.5,3.5) node[
  scale=0.6,
  text=white,
  rotate=0.0
]{116};
\draw (axis cs:5.5,3.5) node[
  scale=0.6,
  text=white,
  rotate=0.0
]{243};
\draw (axis cs:6.5,3.5) node[
  scale=0.6,
  text=white,
  rotate=0.0
]{164};
\draw (axis cs:7.5,3.5) node[
  scale=0.6,
  text=white,
  rotate=0.0
]{33};
\draw (axis cs:8.5,3.5) node[
  scale=0.6,
  text=white,
  rotate=0.0
]{18};
\draw (axis cs:9.5,3.5) node[
  scale=0.6,
  text=white,
  rotate=0.0
]{32};
\draw (axis cs:0.5,4.5) node[
  scale=0.6,
  text=white,
  rotate=0.0
]{25};
\draw (axis cs:1.5,4.5) node[
  scale=0.6,
  text=white,
  rotate=0.0
]{1};
\draw (axis cs:2.5,4.5) node[
  scale=0.6,
  text=white,
  rotate=0.0
]{152};
\draw (axis cs:3.5,4.5) node[
  scale=0.6,
  text=white,
  rotate=0.0
]{90};
\draw (axis cs:4.5,4.5) node[
  scale=0.6,
  text=white,
  rotate=0.0
]{0};
\draw (axis cs:5.5,4.5) node[
  scale=0.6,
  text=white,
  rotate=0.0
]{42};
\draw (axis cs:6.5,4.5) node[
  scale=0.6,
  text=white,
  rotate=0.0
]{200};
\draw (axis cs:7.5,4.5) node[
  scale=0.6,
  text=white,
  rotate=0.0
]{110};
\draw (axis cs:8.5,4.5) node[
  scale=0.6,
  text=white,
  rotate=0.0
]{22};
\draw (axis cs:9.5,4.5) node[
  scale=0.6,
  text=white,
  rotate=0.0
]{9};
\draw (axis cs:0.5,5.5) node[
  scale=0.6,
  text=white,
  rotate=0.0
]{12};
\draw (axis cs:1.5,5.5) node[
  scale=0.6,
  text=white,
  rotate=0.0
]{2};
\draw (axis cs:2.5,5.5) node[
  scale=0.6,
  text=white,
  rotate=0.0
]{64};
\draw (axis cs:3.5,5.5) node[
  scale=0.6,
  text=white,
  rotate=0.0
]{243};
\draw (axis cs:4.5,5.5) node[
  scale=0.6,
  text=white,
  rotate=0.0
]{77};
\draw (axis cs:5.5,5.5) node[
  scale=0.6,
  text=white,
  rotate=0.0
]{0};
\draw (axis cs:6.5,5.5) node[
  scale=0.6,
  text=white,
  rotate=0.0
]{78};
\draw (axis cs:7.5,5.5) node[
  scale=0.6,
  text=white,
  rotate=0.0
]{53};
\draw (axis cs:8.5,5.5) node[
  scale=0.6,
  text=white,
  rotate=0.0
]{10};
\draw (axis cs:9.5,5.5) node[
  scale=0.6,
  text=white,
  rotate=0.0
]{10};
\draw (axis cs:0.5,6.5) node[
  scale=0.6,
  text=white,
  rotate=0.0
]{9};
\draw (axis cs:1.5,6.5) node[
  scale=0.6,
  text=white,
  rotate=0.0
]{4};
\draw (axis cs:2.5,6.5) node[
  scale=0.6,
  text=white,
  rotate=0.0
]{72};
\draw (axis cs:3.5,6.5) node[
  scale=0.6,
  text=white,
  rotate=0.0
]{94};
\draw (axis cs:4.5,6.5) node[
  scale=0.6,
  text=white,
  rotate=0.0
]{160};
\draw (axis cs:5.5,6.5) node[
  scale=0.6,
  text=white,
  rotate=0.0
]{34};
\draw (axis cs:6.5,6.5) node[
  scale=0.6,
  text=white,
  rotate=0.0
]{0};
\draw (axis cs:7.5,6.5) node[
  scale=0.6,
  text=white,
  rotate=0.0
]{7};
\draw (axis cs:8.5,6.5) node[
  scale=0.6,
  text=white,
  rotate=0.0
]{11};
\draw (axis cs:9.5,6.5) node[
  scale=0.6,
  text=white,
  rotate=0.0
]{12};
\draw (axis cs:0.5,7.5) node[
  scale=0.6,
  text=white,
  rotate=0.0
]{16};
\draw (axis cs:1.5,7.5) node[
  scale=0.6,
  text=white,
  rotate=0.0
]{1};
\draw (axis cs:2.5,7.5) node[
  scale=0.6,
  text=white,
  rotate=0.0
]{30};
\draw (axis cs:3.5,7.5) node[
  scale=0.6,
  text=white,
  rotate=0.0
]{53};
\draw (axis cs:4.5,7.5) node[
  scale=0.6,
  text=white,
  rotate=0.0
]{107};
\draw (axis cs:5.5,7.5) node[
  scale=0.6,
  text=white,
  rotate=0.0
]{54};
\draw (axis cs:6.5,7.5) node[
  scale=0.6,
  text=white,
  rotate=0.0
]{22};
\draw (axis cs:7.5,7.5) node[
  scale=0.6,
  text=white,
  rotate=0.0
]{0};
\draw (axis cs:8.5,7.5) node[
  scale=0.6,
  text=white,
  rotate=0.0
]{11};
\draw (axis cs:9.5,7.5) node[
  scale=0.6,
  text=white,
  rotate=0.0
]{23};
\draw (axis cs:0.5,8.5) node[
  scale=0.6,
  text=white,
  rotate=0.0
]{105};
\draw (axis cs:1.5,8.5) node[
  scale=0.6,
  text=white,
  rotate=0.0
]{38};
\draw (axis cs:2.5,8.5) node[
  scale=0.6,
  text=white,
  rotate=0.0
]{22};
\draw (axis cs:3.5,8.5) node[
  scale=0.6,
  text=white,
  rotate=0.0
]{19};
\draw (axis cs:4.5,8.5) node[
  scale=0.6,
  text=white,
  rotate=0.0
]{15};
\draw (axis cs:5.5,8.5) node[
  scale=0.6,
  text=white,
  rotate=0.0
]{7};
\draw (axis cs:6.5,8.5) node[
  scale=0.6,
  text=white,
  rotate=0.0
]{16};
\draw (axis cs:7.5,8.5) node[
  scale=0.6,
  text=white,
  rotate=0.0
]{5};
\draw (axis cs:8.5,8.5) node[
  scale=0.6,
  text=white,
  rotate=0.0
]{0};
\draw (axis cs:9.5,8.5) node[
  scale=0.6,
  text=white,
  rotate=0.0
]{52};
\draw (axis cs:0.5,9.5) node[
  scale=0.6,
  text=white,
  rotate=0.0
]{32};
\draw (axis cs:1.5,9.5) node[
  scale=0.6,
  text=white,
  rotate=0.0
]{126};
\draw (axis cs:2.5,9.5) node[
  scale=0.6,
  text=white,
  rotate=0.0
]{11};
\draw (axis cs:3.5,9.5) node[
  scale=0.6,
  text=white,
  rotate=0.0
]{23};
\draw (axis cs:4.5,9.5) node[
  scale=0.6,
  text=white,
  rotate=0.0
]{6};
\draw (axis cs:5.5,9.5) node[
  scale=0.6,
  text=white,
  rotate=0.0
]{8};
\draw (axis cs:6.5,9.5) node[
  scale=0.6,
  text=white,
  rotate=0.0
]{15};
\draw (axis cs:7.5,9.5) node[
  scale=0.6,
  text=white,
  rotate=0.0
]{14};
\draw (axis cs:8.5,9.5) node[
  scale=0.6,
  text=white,
  rotate=0.0
]{60};
\draw (axis cs:9.5,9.5) node[
  scale=0.6,
  text=white,
  rotate=0.0
]{0};
\end{axis}

\end{tikzpicture}
        \caption{Only misclassifications}
        \label{fig:confusion_all_models_a}
    \end{subfigure}
    \begin{subfigure}[b]{0.42\textwidth}
        \centering
        \input{Images/ConfusionMatrix.tex}
        \caption{All predictions}
        \label{fig:confusion_all_models_b}
    \end{subfigure}
    \caption{Averaged confusion matrices of all models for adversarially perturbed inputs. Subfigure (a) shows the confusion matrix for only successful attacks while subfigure (b) shows all predictions. Both matrices are calculated on the test set of CIFAR10.}
    \label{fig:confusion_all_models}
\end{figure}

%Confusion matrix of misclassifications
We additionally explored the average of the confusion matrices of all models for adversarially perturbed images. Note that the CIFAR10 dataset is balanced and contains an equal amount of samples for all classes. Figure \ref{fig:confusion_all_models_a} shows the confusion matrix of only the misclassifications, while Figure \ref{fig:confusion_all_models_b} shows the whole confusion matrix. It can be seen that the matrix in Figure \ref{fig:confusion_all_models_a} contains only a few large values, which is in line with the previous observation that some classes are easier to perturb than others. Furthermore, the matrix is largely symmetric. Classes are mainly confused amongst pairs. This includes semantically meaningful pairs such as "cats" and "dogs" or "car" and "truck", but it also includes other pairs that generally share similar image backgrounds such as "plane" and "ship", "deer" and "frog", and "deer" and "bird". Examples of robust images and non-robust images are included in the appendix.

\subsection{Model properties} \label{sec:model_prop}

\begin{figure}
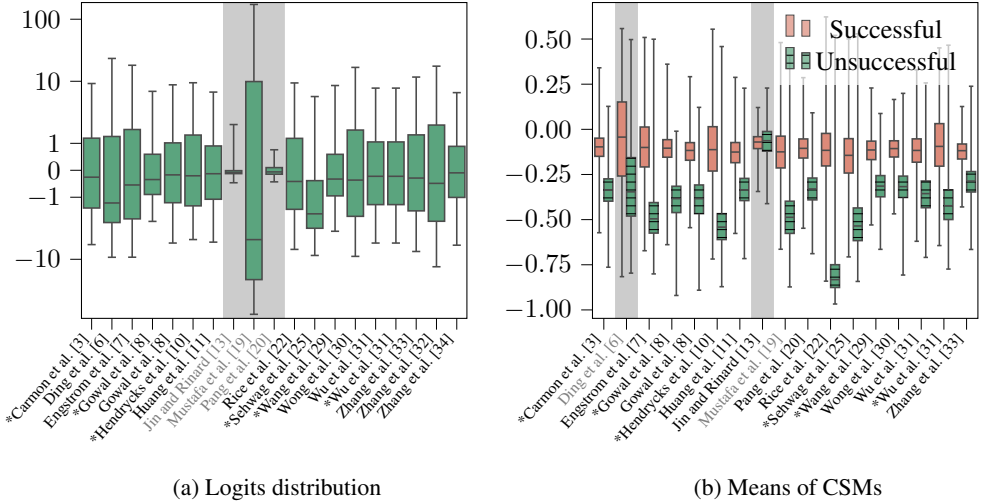

    \centering
    \begin{subfigure}[b]{0.48\textwidth}
        \centering
        \input{Images/Logits_Clean_log}
        \caption{Logits distribution}
        \label{fig:logits_a}
    \end{subfigure}
    \begin{subfigure}[b]{0.48\textwidth}
        \centering
        \input{Images/CSM_Mean_Predicted_Row_BoxPlot}
        \caption{Means of CSMs}
        \label{fig:logits_b}
    \end{subfigure}
    \caption{(a) Box plots of the logits distribution of clean images from the CIFAR10 test set for all models analyzed in this paper. (b) Box plots of the means of the CSMs calculated with the GGA method for both robust (green, dashed) and non-robust (red) images \cite{Schwinn2021Identifying}. The models highlighted by gray shading and text show considerably lower robustness against strong attacks compared to standard \ac{PGD} \cite{Ding2020MMA, Jin2020Manifold, Mustafa2019Adversarial, Pang2020Boosting}. Models that use additional data during training are marked with a *.}
    %Logit values that lead to an overflow in the softmax functions are indicated with a dashed line.
    \label{fig:logits}
\end{figure}

Here, we first analyze the distribution of the output logits $z$ and subsequently inspect the input gradient geometry of the different models \cite{Schwinn2021Identifying}. We then relate these properties to the difficulty of the robustness evaluation. Models that display irregular properties are highlighted with gray shading and text in Figure \ref{fig:logits}.
%Apart from the classification decisions of the models, we inspect the distribution of their output logits $z$.

Prior work observed that simply scaling the output of a \ac{DNN} will lead to vanishing gradients when the softmax function is used in the last layer of the network \cite{croce2020reliable, Gupta2020Improved}. This phenomenon occurs due to finite arithmetic and thus limited precision, where the \ac{CE} loss is quantized to $0$ and the model effectively obfuscates the gradient from the attack. The \ac{CE} loss is given by: 
\begin{equation}
    \operatorname{CE}(z, y) = -\log(\operatorname{softmax(z)_y}) \text{ where } \operatorname{softmax}(z)_y = \frac{e^{z_y}}{\sum_{j=1}^{C} e^{z_j}}.
\end{equation}

Figure \ref{fig:logits_a} summarizes the logit distributions of all models. The model proposed by \citet{Mustafa2019Adversarial} shows large logits, which lead to precision issues as described above. Furthermore, the models by \citet{Mustafa2019Adversarial} and \citet{Ding2020MMA} exhibit a considerably higher average confidence ($0.948$) than all other models ($0.666$ excluding models with exceptionally low confidence \cite{Jin2020Manifold, Pang2020Boosting}). In contrast, the models by \citet{Jin2020Manifold} and \citet{Pang2020Boosting} reveal a different phenomenon where the logits are close to zero and show a substantially lower standard deviation than the other models. Consequently, the logits are generally mapped to a limited value range by the softmax function, where all values are similar. Thus, the loss may only change slightly between different attack iterations, which subsequently decreases the attack performance. This is also reflected in low confidence for the two models, where the most confident prediction has a probability of $<0.51$ while it is $\approx 1$ for all other models. The confidence distribution for all models is given in the appendix.

We further analyzed if robust and non-robust images exhibit different properties with the Geometric Gradient Analysis (GGA) method proposed by \citet{Schwinn2021Identifying}. GGA is used to identify untrustworthy predictions (e.g., adversarial examples) by analyzing the geometry of the saliency maps for a given sample through the computation of Cosine Similarity Matrices (CSMs). In this experiment, we analyzed the clean version of the images. Figure \ref{fig:logits_b} shows a considerable difference between the distribution of the mean values of the CSMs for non-robust images (red) and robust images (green, dashed). This indicates a different response of the \acp{DNN} for non-robust and robust images even if the images are not perturbed. Furthermore, the two models proposed by \citet{Ding2020MMA} and \citet{Pang2020Boosting} (gray shaded area) display a different behavior for the CSM mean values. For those models, the mean values between robust and non-robust images differ only slightly and the box plots overlap while they are substantially different for the other models. 

All $4$ models identified in the above analysis show high robustness against \ac{CE}-based attacks. However, for these $4$ models, the difference between a standard robustness evaluation with \ac{CE}-based \ac{PGD} and stronger attacks is larger than $7\%$ and considerably less accurate than for the other $15$ models \cite{croce2020robustbench}.

%The four models proposed by \citet{Ding2020MMA}, \citet{Jin2020Manifold}, \citet{Mustafa2019Adversarial}, and \citet{Pang2020Boosting} (gray shaded areas in Figures \ref{fig:logits_a} and \ref{fig:logits_b}) show high robustness against softmax-\ac{CE}-based attacks. However, the difference between a standard robustness evaluation with \ac{CE}-based \ac{PGD} and stronger attacks is larger than $7\%$ for those three models \cite{croce2020robustbench}.

%It is possible to identify all models that are difficult to attack by looking at both the output space distribution and the geometry of the input gradients.  
%Additionally, the model proposed by \citet{Mustafa2019Adversarial} and \citet{Pang2020Boosting} also show a different behaviour for the mean values of the CSMs.

\section{Enhancing adversarial attacks}

In the previous section, we explored the misclassification of robust \acp{DNN} under adversarial attacks. The experiments showed a general consistency between the different models. Specifically, we discovered that common attacks mostly focus on a limited amount of different classes to attack in the untargeted setting. Additionally, we observed that the scale and distribution of the output logits are linked to the success rate of adversarial attacks. Based on these observations, we now design a novel loss function for adversarial attacks to make them more effective. We first describe the two main components of this loss function. Subsequently, we elaborate how we can minimize the norm of the final adversarial perturbation without compromising the attack's success rate. This is important as adversarial attacks should not change the label for human perception, which is linked to the perturbation magnitude.

\paragraph{Scale invariance:} Previous work already demonstrated that high output logits can lead to gradient obfuscation and weaken adversarial attacks \cite{Carlini017Towards,croce2020reliable}. We additionally observe that a small value range of the logits can also lead to attack failure. We propose to scale the softmax function by the following rule:  

\begin{equation} \label{eq:scale}
    \hat{z} = \operatorname{softmax}\left(\alpha \cdot \frac{z}{\operatorname{||z||_\infty}}\right)
\end{equation}

where $\alpha$ is an easy-to-tune scalar value that controls the lowest and largest possible output values of the softmax function.

\paragraph{Encouraging diverse attack targets:}

Figure \ref{fig:confusion_all_models} demonstrates that untargeted adversarial attacks mainly induce misclassifications for a limited amount of classes. We argue that this behavior limits the effectiveness of adversarial attacks. This notion is supported by prior work that showed that performing targeted attacks against every possible class is usually more effective than applying a single untargeted attack \cite{Kwon2018Multi,croce2020reliable}. However, these so-called multi-targeted attacks are computationally expensive and do not scale to datasets with a high number of output classes. We propose to exchange the \ac{CE} loss function with the Euclidean distance between the rescaled softmax output $\hat{z}$ and the one-hot encoded ground truth vector $Y$. The Euclidean distance increases fastest by maximizing the magnitude of any of the output logits where $z_i \neq z_y$, which simultaneously minimizes the distance between $Y$ and $z_y$. Combining the euclidean loss function and the scaling described in \eqref{eq:scale} the loss function can be described by the following equation.

\begin{equation} \label{eq:l2_loss}
    \mathcal{L}_2 = ||\hat{z} - Y||_2.
\end{equation}

To encourage the attack to explore different gradient directions we additionally perturb the logits after each forward pass with Gaussian noise, where the noise magnitude is controlled by the hyperparameter $\sigma$. The resulting loss function is given by:

\begin{equation}\label{eq:lossfunc}
    \mathcal{L}_{Noise} = ||\hat{z} + \mathcal{N}(0, \sigma) - Y||_2.
\end{equation}

Note that this method does not improve the performance when using the \ac{CE} loss in our experiments. This is expected as the \ac{CE} loss is only dependent on the output of the ground truth class and adding noise to the other output values has no impact.  

\begin{table}
    \caption{Ablation results for the individual \LossFunc{} components for the model proposed by \citet{Jin2020Manifold}}
    \label{tab:ablation_results}
    \centering
    \begin{tabular}{lrrr}
        \toprule
         Attack & Accuracy & Improvement \\
         \midrule
         APGD\textsubscript{CE} & 52.34 & N/A \\
         APGD\textsubscript{CE \& Scaled} & 18.29 & +34.05 \\
         APGD\textsubscript{Scaled \& L2} & 18.13 & +0.16 \\
         APGD\textsubscript{Scaled \& L2 \& Jitter} & 7.54 & +10.59 \\
        \bottomrule
    \end{tabular}
\end{table}

\paragraph{Minimizing the norm of the perturbation}

Finally, we aim to encourage the attack to find small perturbations. As long as no successful perturbation is found, we apply the loss function presented in \eqref{eq:lossfunc}. Once the adversarial attack is successful we additionally consider the norm of the adversarial perturbation. Furthermore, we only override the current perturbation if the newly found perturbation also leads to a successful attack. This procedure can never decrease the success rate of the attack and effectively minimizes the norm of the adversarial perturbation in our experiments. Moreover, the norm (or other distance measure) can be freely chosen according to the respective problem (e.g., $\ell_1$, $\ell_2$, $\ell_{\infty}$) as long as it is differentiable. The final loss function can be described as follows:

\begin{equation}
    \mathcal{L}_{Jitter}  = 
    \begin{cases}
    \frac{||\hat{z} - Y + \mathcal{N}(0, \sigma)||_2}{||\gamma||_p} & \text{if $x_{adv}$ is misclassified} \\
    ||\hat{z} - Y + \mathcal{N}(0, \sigma)||_2 & \text{if $x_{adv}$ is not misclassified yet}
    \end{cases}.
\end{equation}

The effect of the different components is exemplified in Table~\ref{tab:ablation_results} for the model proposed in \cite{Jin2020Manifold}. Every component decreases the accuracy of the model and therefore increases the success rate of the attack. Note that the norm minimization does not affect the performance and is excluded.

\section{Experiments} \label{sec:experiments}

We conducted a series of experiments to evaluate the effectiveness of the proposed \LossFunc{} loss function. Furthermore, we inspect the perturbations generated with the \LossFunc{} loss function to explain its effectiveness compared to other state-of-the-art loss functions. 

\paragraph{Data and models}

All experiments were performed on the CIFAR10 dataset \cite{Krizhevsky2009}. We chose CIFAR10 as it is one of the most often used datasets to evaluate adversarial robustness. We gathered $\modelCount$ models from the literature for the attack evaluation. All models are either taken from the RobustBench library \cite{croce2020robustbench} or from the GitHub repositories of the authors directly \cite{Mustafa2019Adversarial, Jin2020Manifold}. We only consider models which are trained to be robust against $\ell_{\infty}$-norm attacks. The resulting benchmark contains a diverse set of models which are trained with different methods.

\paragraph{Threat model} \label{sec:threat_model}

\begin{table}
    \small
    \caption{Accuracy of the evaluated models when attacked with APGD using different loss functions. The difference between the best and second-best loss function is given in the right-most column. The most successful attack is highlighted in bold while the second most successful attack is underlined and models that use additional data are marked with a *.}
    \label{tab:lossfunc_attack}
    \centering
    \begin{tabular}{lrrrrr}
        \toprule
        Model & CE & \ac{CW} & DLR &  \LossFunc{} & Diff. \\ 
        \hline
\citet{Mustafa2019Adversarial} & 19.12 & 0.10 & \underline{0.05} & \textbf{0.02} & -0.03 \\
\citet{Jin2020Manifold} & 52.33 & 47.78 & \underline{21.33} & \textbf{7.54} & -13.8 \\
\citet{Wong2020Fast} & \underline{45.83} & 45.95 & 47.05 & \textbf{44.49} & -1.34 \\
\citet{Zhang2019You} & 46.12 & 47.15 & 47.71 & \textbf{46.01} & -0.11 \\
\citet{Ding2020MMA} & \underline{50.13} & 51.07 & 51.29 & \textbf{47.85} & -2.29 \\
\citet{Engstrom2019Robustness} & \underline{51.77} & 52.27 & 53.09 & \textbf{51.08} & -0.69 \\
\citet{Zhang2019Theoretically} & 54.80 & \underline{53.53} & 53.64 & \textbf{53.05} & -0.48 \\
\citet{Huang2020Self} & 55.86 & \underline{53.94} & 54.41 & \textbf{53.33} & -0.60 \\
\citet{Zhang2020Attacks} & 56.84 & \underline{54.49} & 54.77 & \textbf{53.98} & -0.51 \\
\citet{Rice2020Overfitting} & 56.89 & \underline{55.36} & 56.00 & \textbf{54.36} & -1.00 \\
\citet{Pang2020Boosting} & 61.62 & \underline{55.44} & 56.28 & \textbf{54.48} & -0.96 \\
*\citet{Hendrycks2019Using} & 57.15 & \underline{56.44} & 57.23 & \textbf{55.94} & -0.50 \\
\citet{Wu2020Adversarial} & 58.80 & \underline{56.76} & 56.82 & \textbf{56.45} & -0.31 \\
\citet{Gowal2020Uncovering} & 59.50 & 57.82 & \underline{57.60} & \textbf{57.09} & -0.51 \\
*\citet{Wang2020Improving} & 61.82 & \underline{58.23} & 58.95 & \textbf{57.58} & -0.64 \\
*\citet{Sehawag2020Hydra} & 59.61 & \underline{58.30} & 58.45 & \textbf{57.66} & -0.65 \\
*\citet{Carmon2019Unlabeld} & 61.73 & \underline{60.61} & 60.88 & \textbf{60.08} & -0.52 \\
*\citet{Wu2020Adversarial} & 63.32 & \underline{60.62} & 60.67 & \textbf{60.44} & -0.18 \\
*\citet{Gowal2020Uncovering} & 65.69 & \underline{63.75} & 63.92 & \textbf{63.31} & -0.45 \\
    \bottomrule
    \end{tabular}
\end{table}

We compare the performance of different loss functions for the \acf{APGD} attack \cite{croce2020reliable}, which consistently beats other iterative gradient-based attacks. Moreover, \ac{APGD} has no hyperparameters such as step size and thus enables a less biased comparison between different loss functions. We compare \LossFunc{} to three different loss functions. First of all, the \ac{CE} loss which is the standard loss function for training supervised \acp{DNN} and is the most often used loss function for gradient-based adversarial attacks. The \ac{CW} loss proposed by \citet{Carlini017Towards} that shows considerably better results compared to \ac{CE} when the model shows high output logits. The \ac{DLR} loss proposed in \cite{croce2020reliable} that was shown get more stable results compared to the \ac{CE} and \ac{CW} loss. All attacks are untargeted $\ell_{\infty}$-norm attacks with $\epsilon=8/255$ and use $100$ attack iterations.

\paragraph{\LossFunc{} Hyperparameter}

Compared to \ac{CE} and \ac{DLR}, \LossFunc{} introduces two additional hyperparameters. The first hyperparameter $\alpha$ rescales the softmax input and directly controls the possible minimum and maximum value of the output logits and the average magnitude of the gradient. Note that values for $\alpha$ close to or greater than $\approx 83$ will result in an overflow of 32-bit float values in the softmax function and thereby to numerical issues (see Section \ref{sec:model_prop}). Thus, we can focus on $0 < \alpha \ll 83$. In a preliminary experiment, we explored different values for $\alpha$ between $2$ and $20$ and observed a stable performance over all values. We chose $\alpha=10$ for all remaining experiments. The second hyperparameter $\sigma$ controls the amount of noise added to the rescaled softmax output $\hat{z}$. We tuned $\sigma$ for every model individually on a batch of $100$ samples by testing values for $\sigma \in \{0, 0.05, 0.1, 0.15, 0.2\}$. Note that tuning $\sigma$ on a small batch for each model introduces only a negligible overhead ($\approx 1\%$ additional runtime).

\section{Results and discussion}

%In this section, we summarize and discuss the results of the experiments.

\subsection{Attack performance}

Table \ref{tab:lossfunc_attack} compares the performance of the different loss functions on the CIFAR10 dataset. The best result for every model is highlighted in bold. The difference between the best and second-best attack is shown in the rightmost column. The proposed \LossFunc{} loss achieves superior performance compared to all other loss functions for all models. For the model proposed in \cite{Jin2020Manifold}, \LossFunc{} achieves a $13.8\%$ higher success rate than the second best loss function and a $44.8\%$ higher success rate than the commonly used \ac{CE} loss. Moreover, the \LossFunc{} loss is the only loss function that is consistently better than the other loss functions. In contrast, the other three loss functions differ in performance for every model. The \ac{CE} loss is better than \ac{CW} and \ac{DLR} in $4$ out of $\modelCount$ cases, the \ac{CW} loss is better than \ac{CE} and \ac{DLR} in $12$ out of $\modelCount$ cases, and the \ac{DLR} loss is better than \ac{CE} and \ac{CW} in $3$ out of $\modelCount$ cases. To evaluate the performance of \LossFunc{} with a higher computational budget we compared \ac{DLR} and \LossFunc{} using $1000$ model evaluations ($5$ restarts and $200$ iterations). While the success rate increased up to $6.51\%$ for \LossFunc{}, the high budget version of \ac{DLR} performed worse than $100$ iteration \LossFunc{} in all cases. An extensive overview is given in the appendix. 

%While none of the other loss functions is consistently better than the others, the \ac{DLR} loss has the least severe failure cases and gives a reasonable approximation of the robustness for all models except the model proposed by \cite{Jin2020Manifold}. However, its still consistently worse than the proposed loss function in all cases and substantially worse in one case.

\subsection{Induced Misclassifications}

{\def\figheight{4cm}
\begin{figure}
    \centering
    \hfill
    \begin{subfigure}[b]{0.3\textwidth}
        \centering
        % This file was created by tikzplotlib v0.9.8.
\begin{tikzpicture}[trim axis left, trim axis right]

\begin{axis}[
tick align=outside,
tick pos=left,
height=\figheight,
x grid style={white!69.0196078431373!black},
xlabel={Adversarial Pred.},
xmin=0, xmax=10,
xtick style={color=black},
xtick={0.5,1.5,2.5,3.5,4.5,5.5,6.5,7.5,8.5,9.5},
xticklabel style={rotate=90.0},
xticklabels={plane,car,bird,cat,deer,dog,frog,horse,ship,truck},
y dir=reverse,
y grid style={white!69.0196078431373!black},
ylabel={Original Pred.},
ymin=0, ymax=10,
ytick style={color=black},
ytick={0.5,1.5,2.5,3.5,4.5,5.5,6.5,7.5,8.5,9.5},
yticklabels={plane,car,bird,cat,deer,dog,frog,horse,ship,truck}
]
\addplot graphics [includegraphics cmd=\pgfimage,xmin=0, xmax=10, ymin=10, ymax=0] {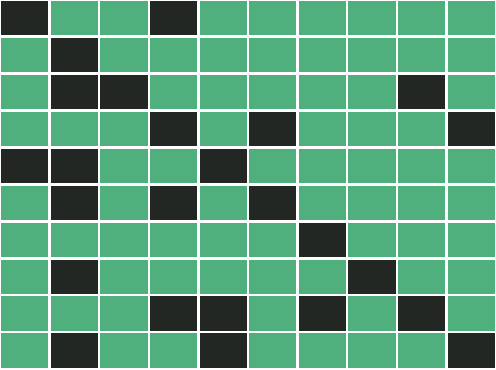};
\end{axis}

\end{tikzpicture}%
        \caption{$\mathcal{L}_{Jitter}$}
        \label{fig:binary_confusion_lossfunc_a}
    \end{subfigure}
    \begin{subfigure}[b]{0.3\textwidth}
        \centering
        % This file was created by tikzplotlib v0.9.8.
\begin{tikzpicture}[trim axis left, trim axis right]

\begin{axis}[
tick align=outside,
tick pos=left,
height=\figheight,
x grid style={white!69.0196078431373!black},
xlabel={Adversarial Pred.},
xmin=0, xmax=10,
xtick style={color=black},
xtick={0.5,1.5,2.5,3.5,4.5,5.5,6.5,7.5,8.5,9.5},
xticklabel style={rotate=90.0},
xticklabels={plane,car,bird,cat,deer,dog,frog,horse,ship,truck},
y dir=reverse,
y grid style={white!69.0196078431373!black},
ymin=0, ymax=10,
ytick style={color=black},
ytick={0.5,1.5,2.5,3.5,4.5,5.5,6.5,7.5,8.5,9.5},
yticklabels={}
]
\addplot graphics [includegraphics cmd=\pgfimage,xmin=0, xmax=10, ymin=10, ymax=0] {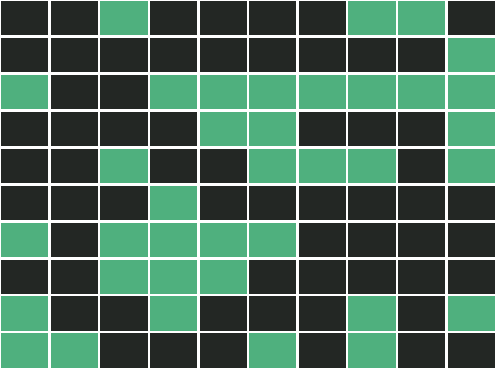};
\end{axis}

\end{tikzpicture}%
        \caption{$\mathcal{L}_2$}
        \label{fig:binary_confusion_lossfunc_b}
    \end{subfigure}
    \begin{subfigure}[b]{0.3\textwidth}
        \centering
        % This file was created by tikzplotlib v0.9.8.
\begin{tikzpicture}[
trim axis left, trim axis right
]

\begin{axis}[
tick align=outside,
tick pos=left,
height=\figheight,
x grid style={white!69.0196078431373!black},
xlabel={Adversarial Pred.},
xmin=0, xmax=10,
xtick style={color=black},
xtick={0.5,1.5,2.5,3.5,4.5,5.5,6.5,7.5,8.5,9.5},
xticklabel style={rotate=90.0},
xticklabels={plane,car,bird,cat,deer,dog,frog,horse,ship,truck},
y dir=reverse,
y grid style={white!69.0196078431373!black},
ymin=0, ymax=10,
ytick style={color=black},
ytick={0.5,1.5,2.5,3.5,4.5,5.5,6.5,7.5,8.5,9.5},
yticklabels={}
]
\addplot graphics [includegraphics cmd=\pgfimage,xmin=0, xmax=10, ymin=10, ymax=0] {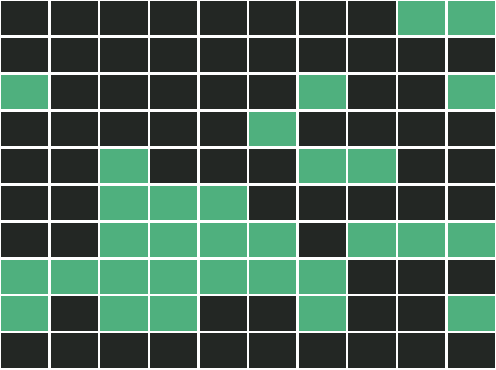};
\end{axis}

\end{tikzpicture}
        \caption{$\mathcal{L}_{DLR}$}
        \label{fig:binary_confusion_lossfunc_c}
    \end{subfigure}
    \hfill
    
    \caption{Illustration of the diversity of targeted classes for $\mathcal{L}_{Jitter}$-based, $\mathcal{L}_2$-based, and $\mathcal{L}_{DLR}$-based attacks. Subfigure (a), (b), and (c) show binarized confusion matrices for the different attacks, where more green squares indicate a higher attack diversity.}
    \label{fig:binary_confusion_lossfunc}
\end{figure}
}

We designed \LossFunc{} to increase the diversity of target classes for untargeted adversarial attacks. Figure \ref{fig:binary_confusion_lossfunc} displays binarized confusion matrices of the model proposed in \cite{Ding2020MMA} for the \ac{APGD} attack. We compare the proposed $\mathcal{L}_{Jitter}$ loss to $\mathcal{L}_{DLR}$ for which we observed the relative sparsity of the confusion matrices in Figure \ref{fig:confusion_all_models_a}. Additionally, we investigate the $\mathcal{L}_2$ loss function given in \eqref{eq:l2_loss} to evaluate the effect of adding Gaussian noise the output logits. We chose the model proposed by \citet{Ding2020MMA}, as both $\mathcal{L}_{DLR}$ and $\mathcal{L}_{2}$ show a considerable performance gap ($>3\%$) compared to $\mathcal{L}_{Jitter}$ for this model.
In the subfigures \ref{fig:binary_confusion_lossfunc_a}, \ref{fig:binary_confusion_lossfunc_b}, and \ref{fig:binary_confusion_lossfunc_c} green squares denote that an attack changed the classification decision to the respective class at least once. $\mathcal{L}_{Jitter}$-based attacks show a considerably higher amount of different target classes compared to the other two attacks. This indicates that adding noise to the logits increases the diversity of an attack. Moreover, $\mathcal{L}_{DLR}$-based attacks were not able to successfully attack the classes car and truck which explains the performance difference to $\mathcal{L}_{Jitter}$.

\begin{figure}
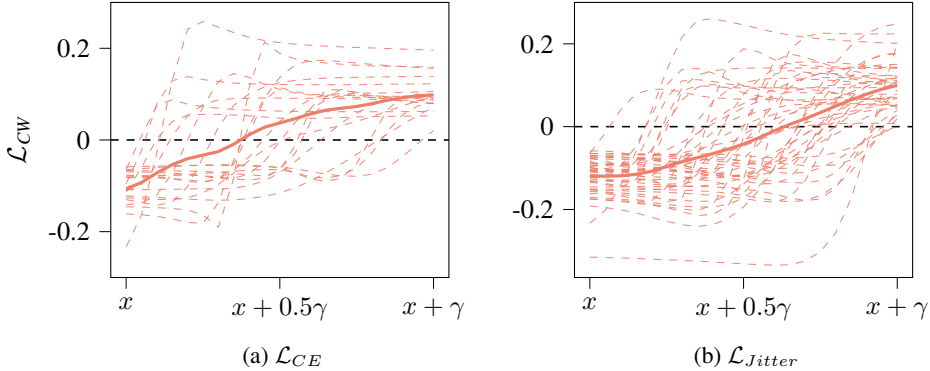

    \centering
    \begin{subfigure}[b]{0.435\textwidth}
        \centering
        \input{Images/ce}
        \caption{$\mathcal{L}_{CE}$}
        \label{loss_comparison_a}
    \end{subfigure}
    \begin{subfigure}[b]{0.435\textwidth}
        \centering
        \input{Images/Jitter}
        \caption{$\mathcal{L}_{Jitter}$}
        \label{loss_comparison_b}
    \end{subfigure}
    %\begin{subfigure}[b]{0.32\textwidth}
    %    \centering
    %    \input{Images/attack_norm}
    %    \caption{L2 norm of attacks}
    %    \label{loss_comparison_c}
    %\end{subfigure}
    \caption{Analysis of the final adversarial perturbation found for $\mathcal{L}_{CE}$-based and $\mathcal{L}_{Jitter}$-based attacks. The \ac{CW} loss \cite{Carlini017Towards} is shown on the y-axis along the direction of an adversarial attack. Here $x$ describes a clean image and $x + \gamma$ the adversarial example.}
    \label{fig:loss_comparison}
\end{figure}

Furthermore, we analyzed the final adversarial perturbations found with \LossFunc{}-based attacks and \ac{CE}-based attacks for the same model \cite{Ding2020MMA}. Figure \ref{fig:loss_comparison} shows the \ac{CW} loss \cite{Carlini017Towards} on the $y$-axis along the direction of an adversarial perturbation on the $x$-axis for both loss functions. We choose the \ac{CW} loss as it can directly be related to the classification decision of a classifier (inputs with $\mathcal{L_{\textit{CW}}} > 0$ are misclassified). The subfigures show the individual loss values for $50$ randomly drawn samples of the test set as dashed lines. The mean value over the whole test set for each group is shown by a solid line. \ac{CE}-based attacks generally find adversarial directions which directly increase the \ac{CW} loss. On the other hand, \LossFunc{}-based attacks mainly find adversarial directions which do not directly increase the \ac{CW} loss, which can be seen by the constant mean near the clean input $x$. Moreover, the mean \ac{CW} loss value of \LossFunc{}-based attacks exceeds the threshold of misclassification noticeably later than \ac{CE}-based attacks (\LossFunc:0.63, CE:0.42). \ac{CE}-based attacks always follow the direction of the steepest ascent. In contrast, \LossFunc{}-based attacks are forced to do more exploration due to the additional noise. This enables \LossFunc{}-based attacks to find perturbation directions that are sub-optimal in the beginning but lead to a misclassification at the final adversarial perturbation.

%\begin{enumerate}
%    \item Does jitter actually lead to more diverse attacks (e.g., more different classes)
%    \item plot loss over attack of targeted and untargeted and jitter as multiple curves
%   \item Does restarting the attack give similar results each time? (If yes argue that its stable, If classes change argue that restarting helps alot ;-))
%    \item models are optimized against untargeted attacks -> robust untargeted attacks
%\end{enumerate}

\subsection{Attack norm and structure}

In a final experiment, we examined the average perturbation norm of the different attack configurations for all $\modelCount$ models. We choose to minimize the $\ell_2$ norm with \LossFunc{}, as differences in the $\ell_2$ norm are easier to interpret than for the $\ell_{\infty}$ norm (e.g. the attack focusing on specific regions). The average $\ell_2$ perturbation norm over all samples for the different loss functions is:  \ac{CE}:$0.52$, \ac{CW}:$0.56$, DLR:$0.55$, and \LossFunc{}:$0.19$. A more extensive overview is given in the appendix. We also inspect the structure of the perturbations. Figure \ref{fig:perturbed_images} displays the perturbation for \ac{CE}- and \LossFunc{}-based attacks for several images. To plot the perturbations, we calculate the absolute sum over every color channel and show the magnitude as a color gradient, where no change is denoted by black color. \ac{CE}-based attacks generally attack every pixel in an image. In comparison, \LossFunc{}-based attacks mainly focus on the salient regions of an image. We argue that focusing on the most distinct image regions enables \LossFunc-based attacks to create successful low-norm adversarial attacks. %Although some of the models show a relatively high robustness against adversarial attacks, they can still be fooled by low norm perturbations in many cases. %This encourages to not only show the robustness for a given $\epsilon$ but further report the average perturbation norm necessary to fool the network to get a more extensive overview about model robustness \cite{Carlini017Towards}. 

%There is a considerable difference between the $\ell_2$ norms of the perturbations. 

%\begin{figure}
%    \centering
    
%    \caption{L2 norm of attacks}
%    \label{fig:perturbed_images}
%\end{figure}

\section{Conclusion and outlook}

In this paper, we analyze the classification decisions of a diverse set of models that are trained to be adversarially robust. We utilize insights of our analysis to create a novel loss function which we name \LossFunc{} that increases the success rate of adversarial attacks. Specifically, we enforce scale invariance of the loss function and encourage a diverse set of target classes for the attack by adding Gaussian noise to the output logits. The proposed method shows superior attack success rates for $19$ out of $19$ models compared to three other popular loss functions in the literature. Moreover, the average perturbation norm of \LossFunc{}-based attacks is considerably lower compared to prior methods, which is achieved without compromising the success rate of the attack. Future work will explore automatically tuning of the noise injection to the output logits for every individual sample during an attack.

\bibliography{bibfile}

\clearpage

\appendix

\section{Appendix}

\subsection{Robust and non-robust images}

\begin{figure}[h]
    \centering
    \begin{subfigure}[b]{0.48\textwidth}
        \centering
        \includegraphics[width=\textwidth]{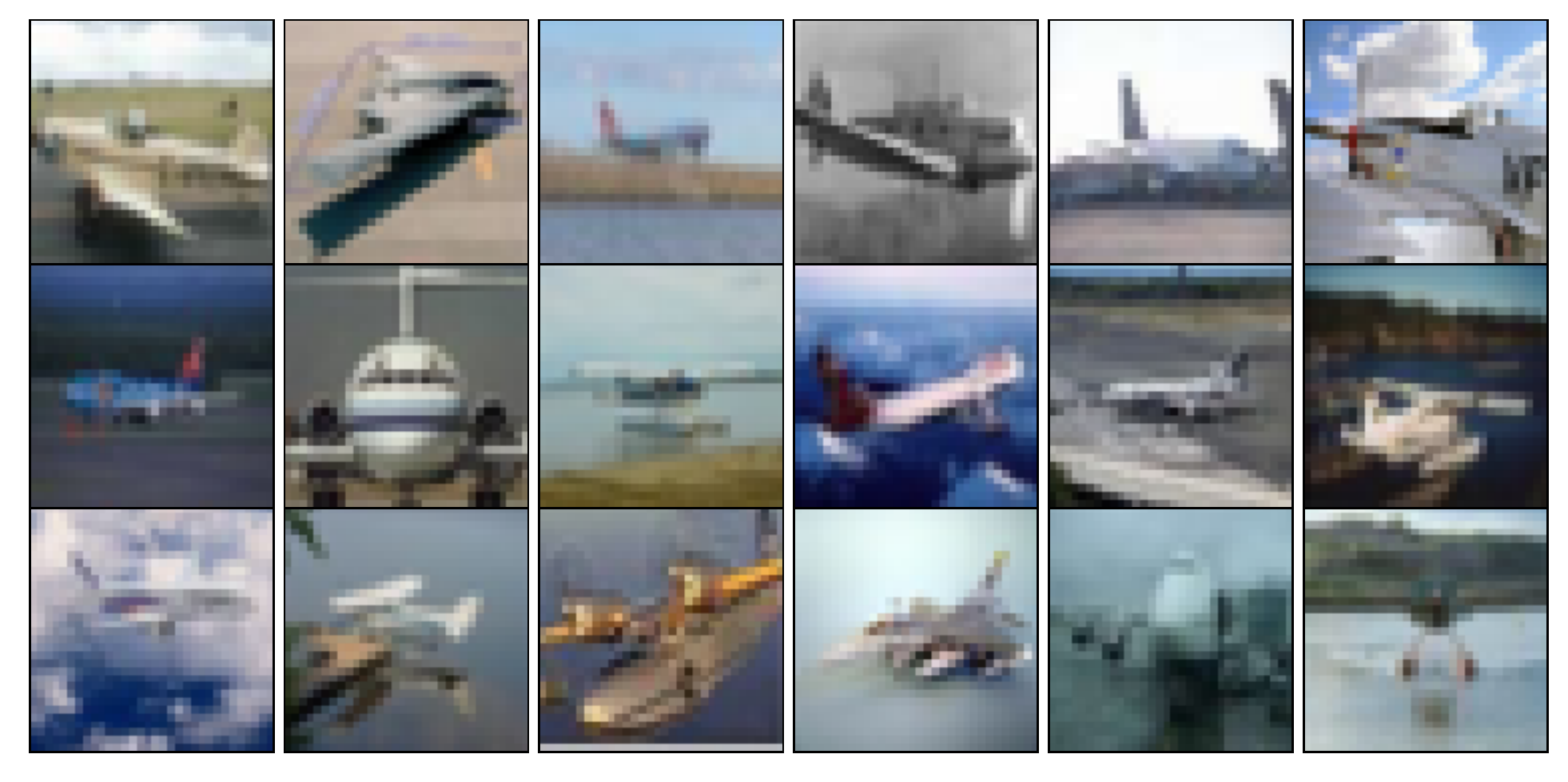}
        \caption{Planes to Ships}
    \end{subfigure}
    \hfill
    \begin{subfigure}[b]{0.48\textwidth}
        \centering
        \includegraphics[width=\textwidth]{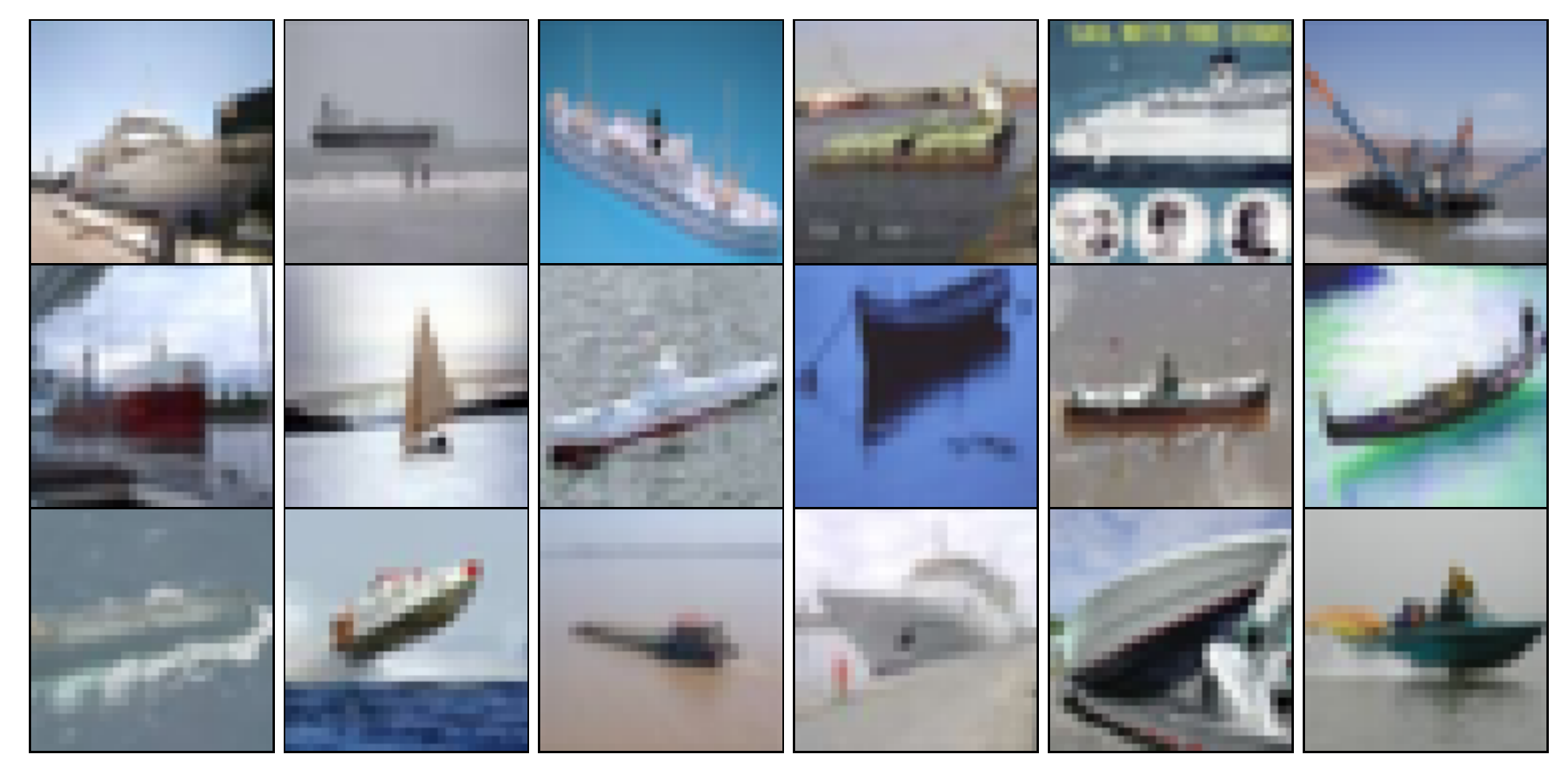}
        \caption{Ships to plane}
    \end{subfigure}
    
    \begin{subfigure}[b]{0.48\textwidth}
        \centering
        \includegraphics[width=\textwidth]{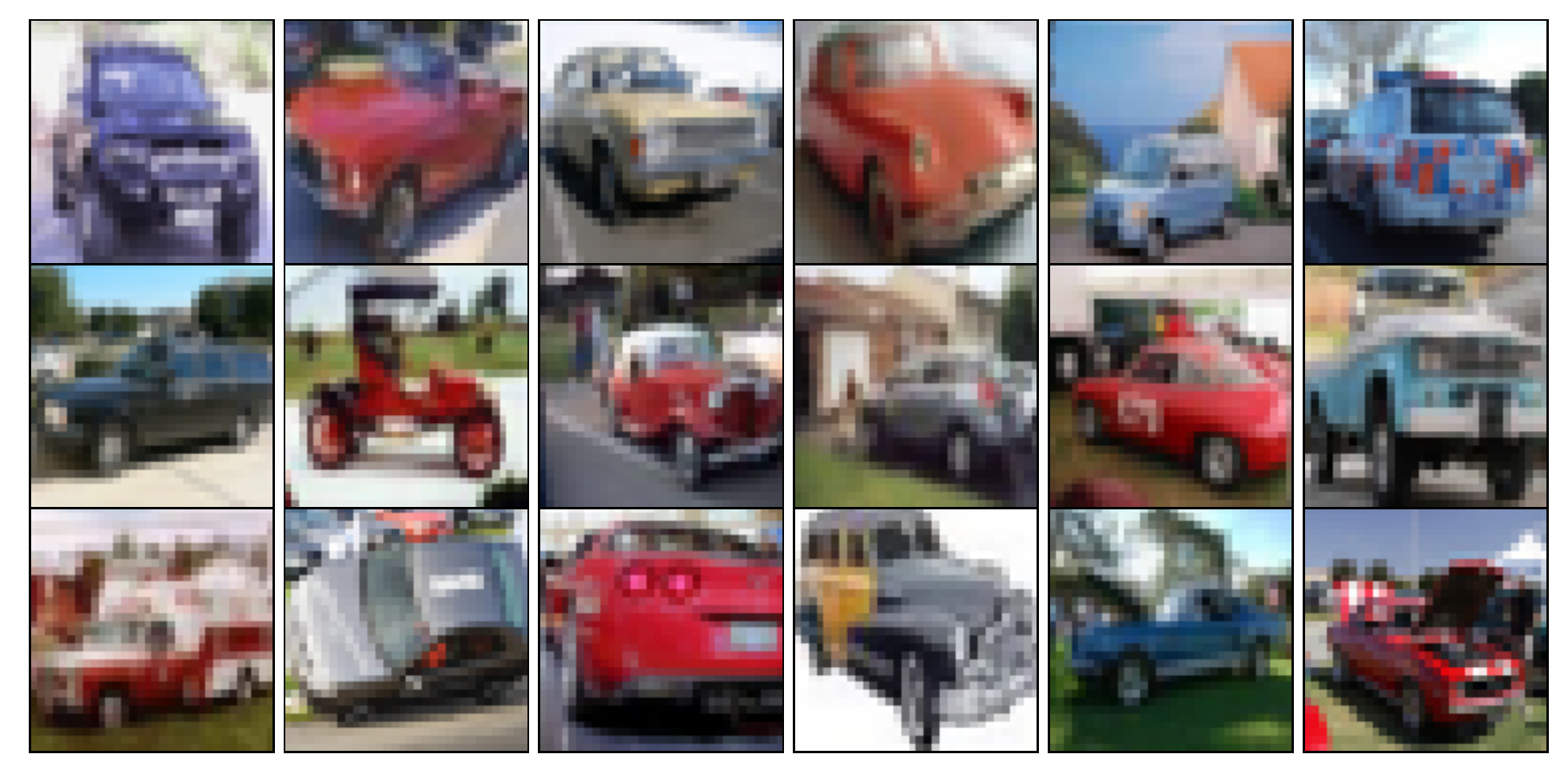}
        \caption{Car to truck}
    \end{subfigure}
    \hfill
    \begin{subfigure}[b]{0.48\textwidth}
        \centering
        \includegraphics[width=\textwidth]{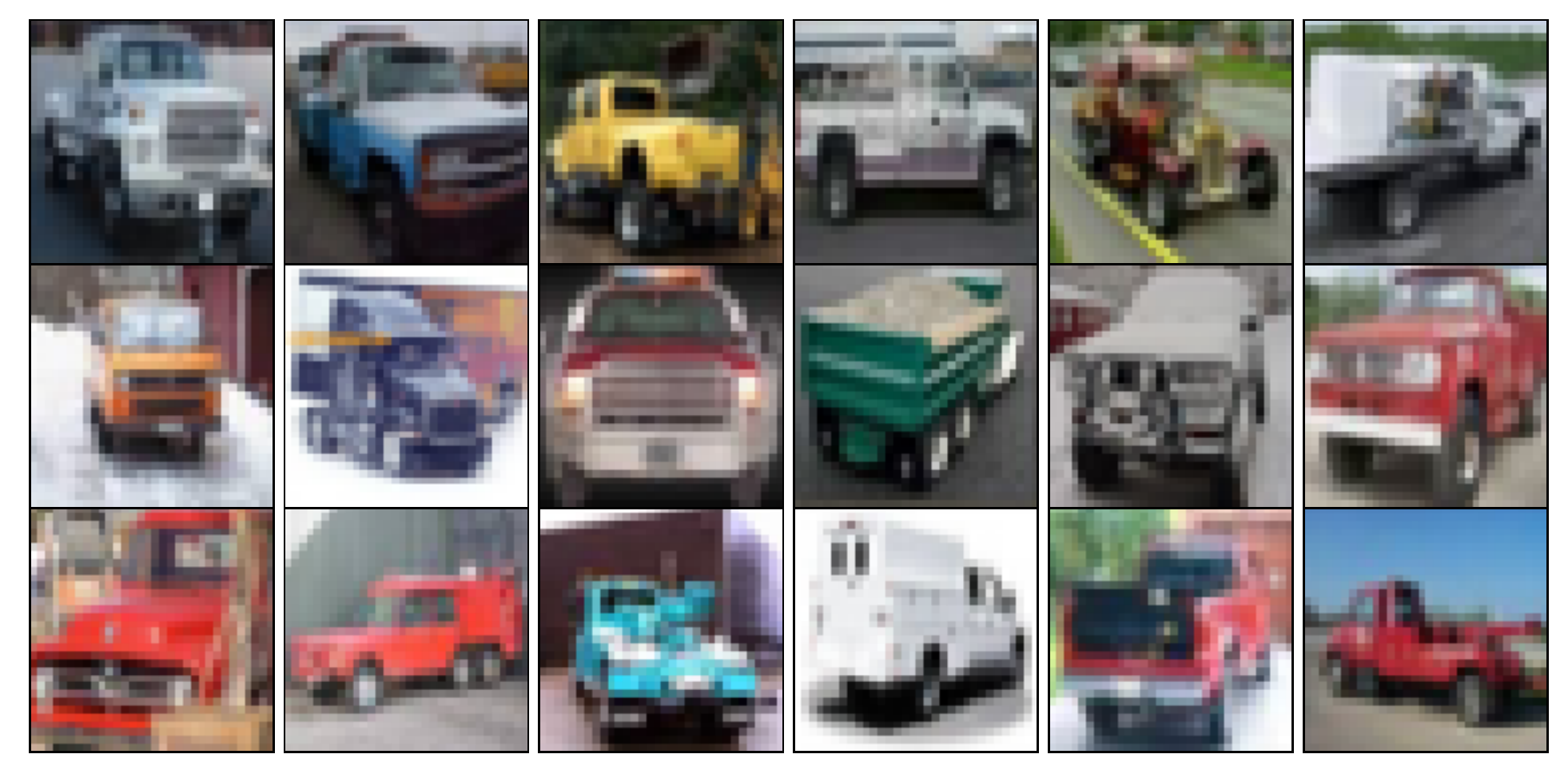}
        \caption{Truck to car}
    \end{subfigure}
    
    \begin{subfigure}[b]{0.48\textwidth}
        \centering
        \includegraphics[width=\textwidth]{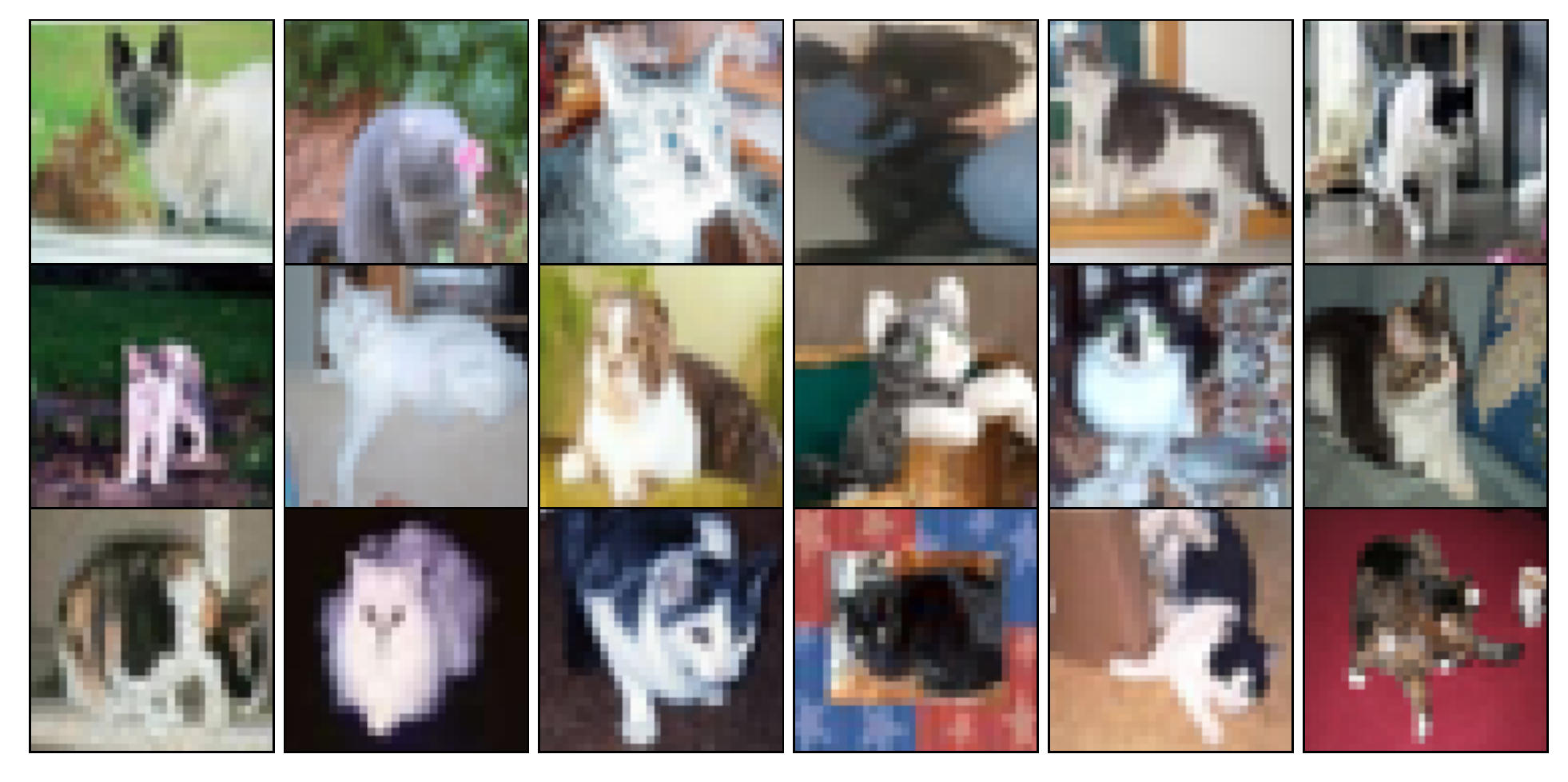}
        \caption{Cat to dog}
    \end{subfigure}
    \hfill
    \begin{subfigure}[b]{0.48\textwidth}
        \centering
        \includegraphics[width=\textwidth]{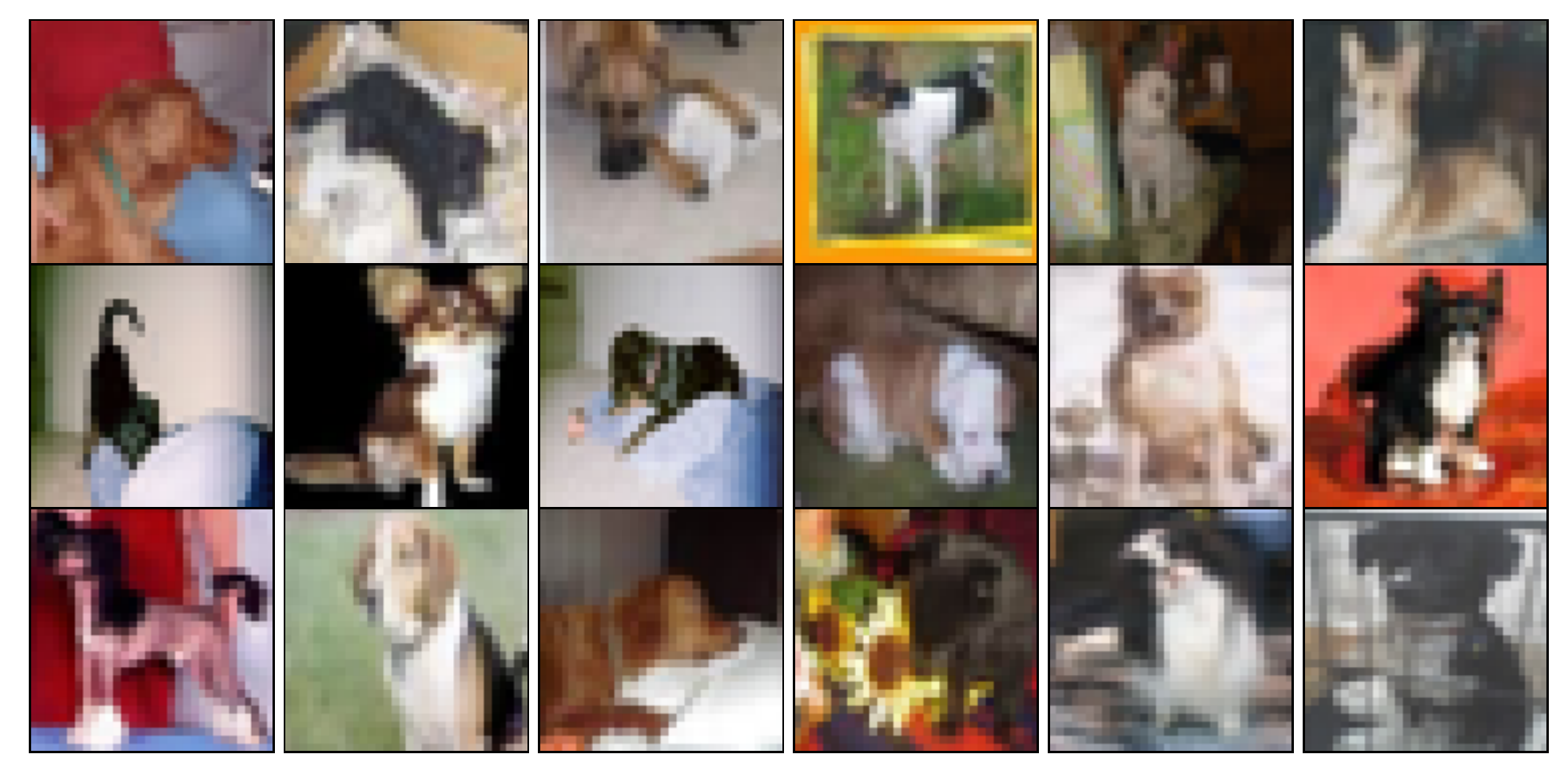}
        \caption{Dog to cat}
    \end{subfigure}
    
    \begin{subfigure}[b]{0.48\textwidth}
        \centering
        \includegraphics[width=\textwidth]{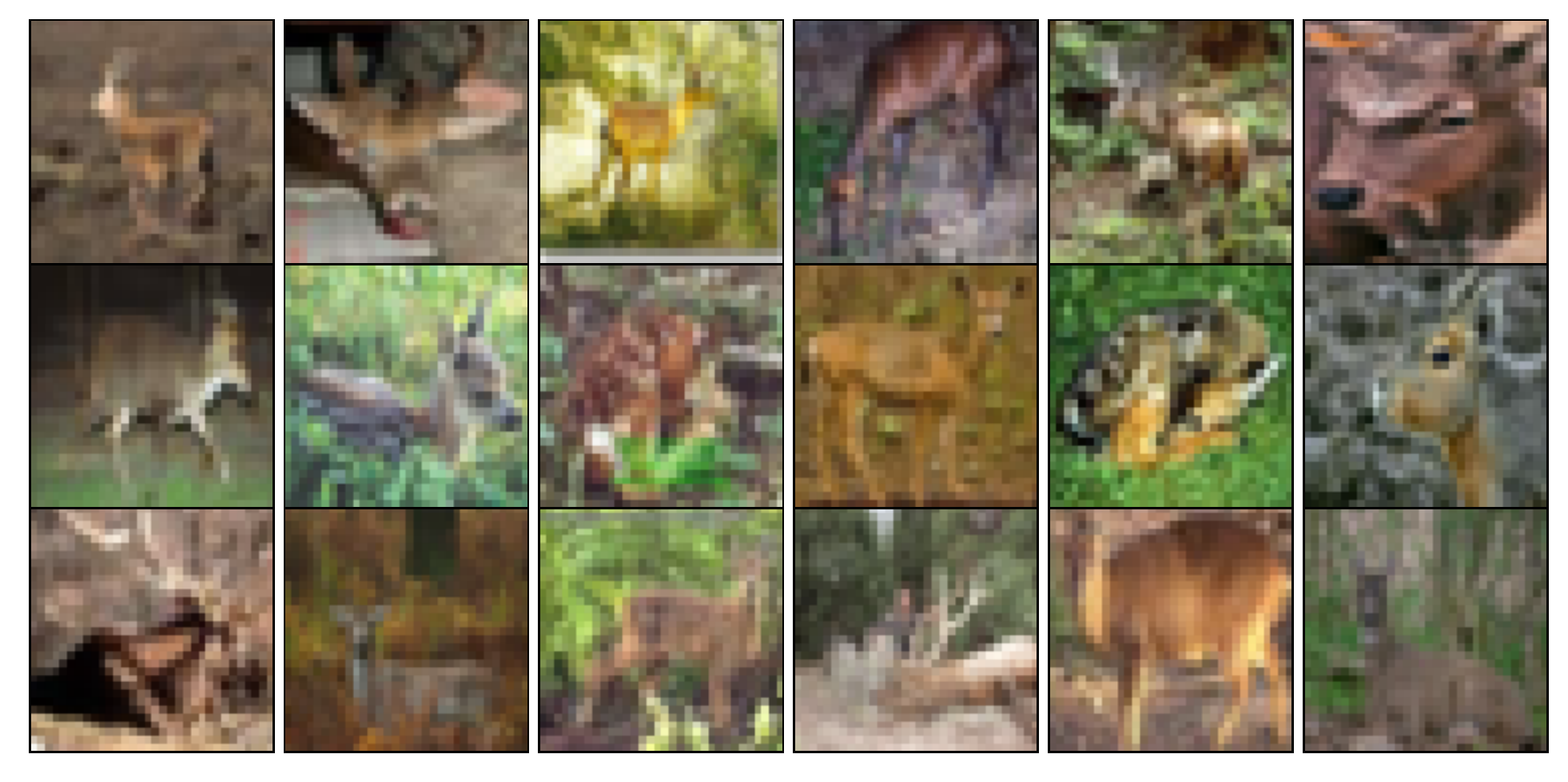}
        \caption{Deer to frog}  
    \end{subfigure}
    \hfill
    \begin{subfigure}[b]{0.48\textwidth}
        \centering
        \includegraphics[width=\textwidth]{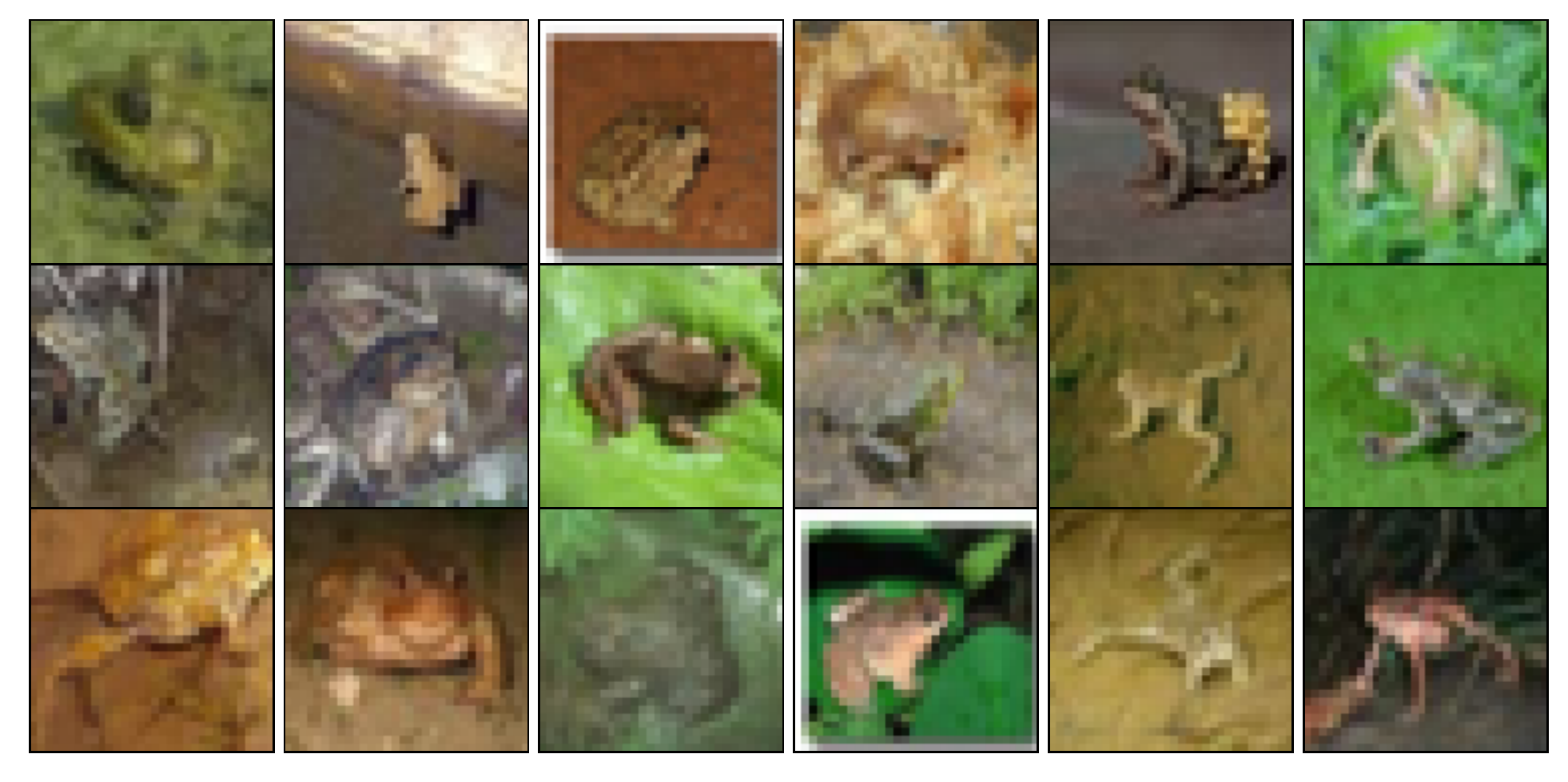}
        \caption{Frog to deer}
    \end{subfigure}
    \caption{Images that are correctly classified by all models under normal conditions but missclassified by all models under attack (DLR-based APGD). Examples of images that are misclassified as the same target class (e.g., cat images that are always misclassified as dogs) are shown.}
    \label{fig:images}
\end{figure}

In Figure \ref{fig:images} we display examples of robust and non-robust images. Specifically, we show images that are correctly classified by all models under normal conditions but are misclassified when attacked. Moreover, we focus on images where all models predict the same wrong target class. These images contain semantically interesting examples:
\begin{itemize}
    \item A ship in the air that is classified as a plane.
    \item A golf cart that is labeled as a car but classified as a truck. 
    \item An ambulance that is labeled as a car but classified as a truck.
\end{itemize}

\subsection{Model confidence distribution}

\begin{figure}[h]
    \centering
    \begin{subfigure}[b]{0.5\textwidth}
        \centering
        \input{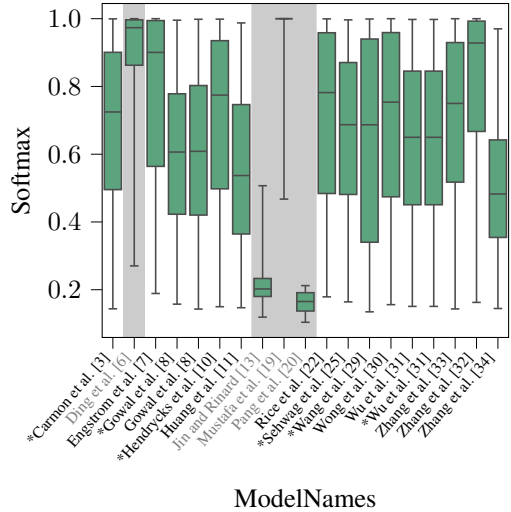}
        \label{fig:softmax_clean}
    \end{subfigure}
    \caption{Confidence distribution of all models. Only the highest softmax output for every prediction is considered.}
    \label{fig:confidence}
\end{figure}

We observed that models that show under- or over-confident predictions on average are more difficult to attack with standard attacks. The confidence distribution for all models is summarized in Figure \ref{fig:confidence}. Models that are either under- or over-confident are highlighted by gray shading and text.

\subsection{Attack norm}

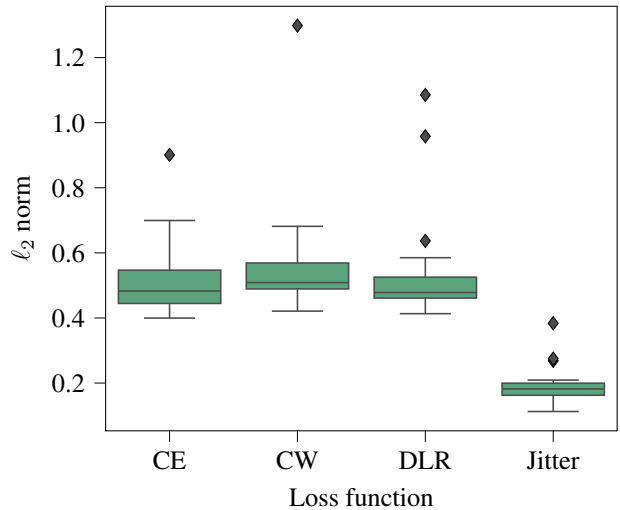
\begin{figure}[h]
    \centering
    \begin{subfigure}[b]{0.6\textwidth}
        \centering
        % This file was created by tikzplotlib v0.9.8.
\begin{tikzpicture}[
trim axis left, trim axis right
]

\definecolor{color0}{rgb}{0.358786434138385,0.641195212753761,0.495720349031661}

\begin{axis}[
tick align=outside,
tick pos=left,
width=\linewidth,
x grid style={white!69.0196078431373!black},
xlabel={Loss function},
xmin=-0.5, xmax=3.5,
xtick style={color=black},
xtick={0,1,2,3},
xticklabels={CE,CW,DLR,Jitter},
y grid style={white!69.0196078431373!black},
ylabel={$\ell_2$ norm},
ymin=0.0532672208070755, ymax=1.35764925458431,
ytick style={color=black},
ytick={0,0.2,0.4,0.6,0.8,1,1.2,1.4},
yticklabels={0.0,0.2,0.4,0.6,0.8,1.0,1.2,1.4}
]
\path [draw=white!29.8039215686275!black, fill=color0, semithick]
(axis cs:-0.4,0.444559504127502)
--(axis cs:0.4,0.444559504127502)
--(axis cs:0.4,0.546925419044495)
--(axis cs:-0.4,0.546925419044495)
--(axis cs:-0.4,0.444559504127502)
--cycle;
\path [draw=white!29.8039215686275!black, fill=color0, semithick]
(axis cs:0.6,0.489254870223999)
--(axis cs:1.4,0.489254870223999)
--(axis cs:1.4,0.568919347190857)
--(axis cs:0.6,0.568919347190857)
--(axis cs:0.6,0.489254870223999)
--cycle;
\path [draw=white!29.8039215686275!black, fill=color0, semithick]
(axis cs:1.6,0.461116461372375)
--(axis cs:2.4,0.461116461372375)
--(axis cs:2.4,0.525453859138489)
--(axis cs:1.6,0.525453859138489)
--(axis cs:1.6,0.461116461372375)
--cycle;
\path [draw=white!29.8039215686275!black, fill=color0, semithick]
(axis cs:2.6,0.162465432262421)
--(axis cs:3.4,0.162465432262421)
--(axis cs:3.4,0.199587696266174)
--(axis cs:2.6,0.199587696266174)
--(axis cs:2.6,0.162465432262421)
--cycle;
\addplot [semithick, white!29.8039215686275!black]
table {%
0 0.444559504127502
0 0.399648503112793
};
\addplot [semithick, white!29.8039215686275!black]
table {%
0 0.546925419044495
0 0.699463145065308
};
\addplot [semithick, white!29.8039215686275!black]
table {%
-0.2 0.399648503112793
0.2 0.399648503112793
};
\addplot [semithick, white!29.8039215686275!black]
table {%
-0.2 0.699463145065308
0.2 0.699463145065308
};
\addplot [black, mark=diamond*, mark size=2.5, mark options={solid,fill=white!29.8039215686275!black}, only marks]
table {%
0 0.900934909820557
};
\addplot [semithick, white!29.8039215686275!black]
table {%
1 0.489254870223999
1 0.421035501670837
};
\addplot [semithick, white!29.8039215686275!black]
table {%
1 0.568919347190857
1 0.681591245269775
};
\addplot [semithick, white!29.8039215686275!black]
table {%
0.8 0.421035501670837
1.2 0.421035501670837
};
\addplot [semithick, white!29.8039215686275!black]
table {%
0.8 0.681591245269775
1.2 0.681591245269775
};
\addplot [black, mark=diamond*, mark size=2.5, mark options={solid,fill=white!29.8039215686275!black}, only marks]
table {%
1 1.29835916213989
};
\addplot [semithick, white!29.8039215686275!black]
table {%
2 0.461116461372375
2 0.413152644348145
};
\addplot [semithick, white!29.8039215686275!black]
table {%
2 0.525453859138489
2 0.585106583023071
};
\addplot [semithick, white!29.8039215686275!black]
table {%
1.8 0.413152644348145
2.2 0.413152644348145
};
\addplot [semithick, white!29.8039215686275!black]
table {%
1.8 0.585106583023071
2.2 0.585106583023071
};
\addplot [black, mark=diamond*, mark size=2.5, mark options={solid,fill=white!29.8039215686275!black}, only marks]
table {%
2 1.08536929473877
2 0.636892449188232
2 0.95806029586792
};
\addplot [semithick, white!29.8039215686275!black]
table {%
3 0.162465432262421
3 0.112557313251495
};
\addplot [semithick, white!29.8039215686275!black]
table {%
3 0.199587696266174
3 0.209097031116486
};
\addplot [semithick, white!29.8039215686275!black]
table {%
2.8 0.112557313251495
3.2 0.112557313251495
};
\addplot [semithick, white!29.8039215686275!black]
table {%
2.8 0.209097031116486
3.2 0.209097031116486
};
\addplot [black, mark=diamond*, mark size=2.5, mark options={solid,fill=white!29.8039215686275!black}, only marks]
table {%
3 0.268394964981079
3 0.274927771186829
3 0.383697702789307
};
\addplot [semithick, white!29.8039215686275!black]
table {%
-0.4 0.482785396575928
0.4 0.482785396575928
};
\addplot [semithick, white!29.8039215686275!black]
table {%
0.6 0.508592855072021
1.4 0.508592855072021
};
\addplot [semithick, white!29.8039215686275!black]
table {%
1.6 0.47816152305603
2.4 0.47816152305603
};
\addplot [semithick, white!29.8039215686275!black]
table {%
2.6 0.181916754436493
3.4 0.181916754436493
};
\end{axis}

\end{tikzpicture}
        \label{loss_comparison_c}
    \end{subfigure}
    \caption{Analysis of the $\ell_2$ norm perturbation magnitude between the different loss functions. The box plots show the quartiles of the data while the whiskers extend to $95\%$ of the value range.}
    \label{fig:l2_norm}
\end{figure}

The distribution of the  $\ell_2$ norm perturbation magnitude is displayed in Figure \ref{fig:l2_norm}. \LossFunc{}-based attacks exhibit perturbations with smaller norms compared to attacks with the other three loss functions in all cases.

\subsection{Attack performance for a higher computational budget}

The performance of DLR- and \LossFunc{}-based attacks for more model evaluations is shown in Table \ref{tab:lossfunc_attack_strong}. Attacks with a "strong" suffix use $200$ iterations and $5$ restarts, while the other attacks use $100$ iterations without additional restarts. Low-budget \LossFunc{}-based attacks achieve a higher success rate than both normal and strong DLR-based attacks in all cases. Overall more model evaluations do only marginally improve the performance for DLR-based attacks except for the model proposed by \citet{Jin2020Manifold}, where the success rate increases by $8.9$ percentage points. For \LossFunc{}-based attacks more model evaluations improve the performance considerably for the models proposed by \citet{Jin2020Manifold} and \citet{Ding2020MMA} and slightly for the models proposed by \citet{Wong2020Fast}, \citet{Rice2020Overfitting}, and \citet{Hendrycks2019Using}. 

\begin{table}[h]
    \small
    \caption{Accuracy [\%] of the evaluated models when attacked with APGD using either the DLR or Jitter loss function. Attacks with a "strong" suffix use $200$ iterations and $5$ restarts, while the other attacks use $100$ iterations without additional restarts.}
    \label{tab:lossfunc_attack_strong}
    \centering
    \begin{tabular}{lrrrrr}
        \toprule
        Models & DLR & Jitter & DLR Strong & Jitter Strong & Diff. \\ 
        \hline
\citet{Mustafa2019Adversarial} & 0.05 & 0.02 & 0.03 & \textbf{0.00} & 0.02 \\
\citet{Jin2020Manifold} & 21.33 & 7.54 & 12.43 & \textbf{1.03} & 6.51 \\
\citet{Wong2020Fast} & 47.05 & 44.49 & 46.69 & \textbf{43.45} & 1.04 \\
\citet{Ding2020MMA} & 51.29 & 47.85 & 50.19 & \textbf{43.62} & 4.23 \\
\citet{Zhang2019You} & 47.71 & 46.01 & 47.31 & \textbf{45.79} & 0.22 \\
\citet{Engstrom2019Robustness} & 53.09 & 51.08 & 52.59 & \textbf{50.83} & 0.24 \\
\citet{Zhang2019Theoretically} & 53.64 & 53.05 & 53.42 & \textbf{52.88} & 0.17 \\
\citet{Huang2020Self} & 54.41 & 53.33 & 54.24 & \textbf{53.25} & 0.09 \\
\citet{Zhang2020Attacks} & 54.77 & 53.98 & 54.54 & \textbf{53.64} & 0.34 \\
\citet{Rice2020Overfitting} & 56.00 & 54.36 & 55.70 & \textbf{53.66} & 0.70 \\
\citet{Pang2020Boosting} & 56.28 & 54.48 & 55.97 & \textbf{54.10} & 0.38 \\
\citet{Hendrycks2019Using} & 57.23 & 55.94 & 56.98 & \textbf{55.10} & 0.84 \\
\citet{Wu2020Adversarial} & 56.82 & 56.45 & 56.69 & \textbf{56.10} & 0.35 \\
\citet{Gowal2020Uncovering} & 57.60 & 57.09 & 57.44 & \textbf{57.08} & 0.01 \\
\citet{Wang2020Improving} & 58.95 & 57.58 & 58.55 & \textbf{57.28} & 0.31 \\
\citet{Sehawag2020Hydra} & 58.45 & 57.66 & 58.23 & \textbf{57.50} & 0.15 \\
\citet{Carmon2019Unlabeld} & 60.88 & 60.08 & 60.62 & \textbf{59.90} & 0.19 \\
\citet{Wu2020Adversarial} & 60.67 & 60.44 & 60.56 & \textbf{60.19} & 0.25 \\
\citet{Gowal2020Uncovering} & 63.92 & 63.31 & 63.74 & \textbf{62.73} & 0.57 \\
    \bottomrule
    \end{tabular}
\end{table}

\subsection{Jitter Code}

The following algorithm shows a PyTorch-like implementation of \LossFunc{}.

\begin{table}[h]
    \centering
    \begin{tabular}{l}
    \toprule
    \textbf{Algoritm 1} Code for the Jitter loss in a PyTorch-like fashion \\
    \hline
    \code{code-green}{\# X:\,~input data, X\_adv:\,~adversarial input data, B:\,~batch size}\\
    \code{code-green}{\# z:\,~logits, y:\,~labels, Y:\,~one-hot encoded labels}\\
    \code{code-green}{\# alpha:\,~value range, sigma:\,~noise magnitude, norm:\,~norm to minimize}\\
    \\
    \code{code-green}{\#\#\#\#\#\#\#\#\#\#\#\#\#\#\#\#\#\#\#\#\#\#\#\#\#\#\#\#\#\# logit scaling \#\#\#\#\#\#\#\#\#\#\#\#\#\#\#\#\#\#\#\#\#\#\#\#\#\#\#}\\
    \code{black}{z\_scaled = z / norm(z.view(B, -1), p=float("inf"), dim=1, keepdim=True)}\\
    \code{black}{z\_scaled = softmax(z\_scaled * alpha, dim=1)}\\
    \code{black}{z\_noisy = z\_scaled + randn\_like(z\_scaled) * sigma} \\
    \code{code-green}{\#\#\#\#\#\#\#\#\#\#\#\#\#\#\#\#\#\#\#\#\#\#\#\#\#\#\#\#\#\# l2 loss \#\#\#\#\#\#\#\#\#\#\#\#\#\#\#\#\#\#\#\#\#\#\#\#\#\#\#\#\#\#\#\#\#}\\ 
    \code{black}{l2 = norm((z\_noisy - Y).view(B, -1), p=2, dim=1)}\\
    \code{code-green}{\#\#\#\#\#\#\#\#\#\#\#\#\#\#\#\#\#\#\#\#\#\#\#\#\#\#\#\#\#\# perturbation magnitude \#\#\#\#\#\#\#\#\#\#\#\#\#\#\#\#\#\#}\\
    \code{black}{non\_adversarial\_mask = z.argmax(1) != y}\\
    \code{black}{magnitude = norm((X - X\_adv).view(B, -1), p=norm, dim=1)}\\
    \code{black}{masked\_magnitude = ones\_like(l2)}\\
    \code{black}{masked\_magnitude[non\_adversarial\_mask] = magnitude[non\_adversarial\_mask]}\\
    \code{code-green}{\#\#\#\#\#\#\#\#\#\#\#\#\#\#\#\#\#\#\#\#\#\#\#\#\#\#\#\#\#\# final loss \#\#\#\#\#\#\#\#\#\#\#\#\#\#\#\#\#\#\#\#\#\#\#\#\#\#\#\#\#\#}\\ 
    \code{black}{loss = l2 / masked\_magnitude}\\
    \code{black}{return loss}\\
    \bottomrule
    \end{tabular}
\end{table}

\end{document}